\documentclass[journal]{IEEEtran}
\usepackage{graphicx}
\usepackage{comment}
\usepackage{amsmath,amssymb} 
\usepackage{color}
\usepackage{multirow}
\usepackage{float}
\usepackage{wrapfig}
\usepackage{graphicx}
\usepackage{enumitem}
\usepackage{animate}
\usepackage{color}
\usepackage{cite}
\usepackage[pagebackref=true,breaklinks=true,colorlinks,bookmarks=false]{hyperref}
\newcommand{\R}{\mathbb{R}}
\definecolor{ao}{rgb}{0.0, 0.5, 0.0}

\newcommand{\revise}[1]{{\color{black} #1}}
\newcommand{\reviseminor}[1]{{\color{black} #1}}

\ifCLASSINFOpdf
\else
\fi
\begin{document}
%
\title{Visual Explanation for Deep Metric Learning}
%
%
%

\author{Sijie Zhu,
        Taojiannan Yang,
        and~Chen~Chen,~\IEEEmembership{Member,~IEEE}
\thanks{S. Zhu, T. Yang, and C. Chen are with the Center for Research in Computer Vision, University of Central Florida, Orlando, FL 32816 USA. E-mails: sizhu@knights.ucf.edu, taoyang1122@knights.ucf.edu, chen.chen@crcv.ucf.edu}
\thanks{This work is partially supported by the National Science Foundation under Grant No. 1910844.}
}

%
%

\markboth{Journal of \LaTeX\ Class Files,~Vol.~14, No.~8, August~2015}%
{Shell \MakeLowercase{\textit{et al.}}: Bare Demo of IEEEtran.cls for IEEE Journals}
%



\maketitle

\begin{abstract}
This work explores the visual explanation for deep metric learning and its applications. As an important problem for learning representation, metric learning has attracted much attention recently, while the interpretation of the metric learning model is not as well-studied as classification. To this end, we propose an intuitive idea to show where contributes the most to the overall similarity of two input images by decomposing the final activation. Instead of only providing the overall activation map of each image, we propose to generate point-to-point activation intensity between two images so that the relationship between different regions is uncovered. We show that the proposed framework can be directly applied to a wide range of metric learning applications and provides valuable information for model understanding. Both theoretical and empirical analyses are provided to demonstrate the superiority of the proposed overall activation map over existing methods. 
Furthermore, our experiments validate the effectiveness of the proposed point-specific activation map on two applications, i.e. cross-view pattern discovery and interactive retrieval. Code is available at \url{https://github.com/Jeff-Zilence/Explain_Metric_Learning}
\end{abstract}


%
\IEEEpeerreviewmaketitle

\section{Introduction}
\label{sec:intro}
%
%
%
%
\IEEEPARstart{L}{earning} the similarity metrics between arbitrary images is a fundamental problem for a variety of tasks, such as image retrieval \cite{Lifted}, verification \cite{FaceNet,Re-ID}, localization \cite{cvmnet}, video tracking \cite{Siamese_fc}, etc. Recently the deep Siamese network \cite{Siamese} based framework has become a standard architecture for metric learning and achieves exciting results on a wide range of applications \cite{MS}. However, there are surprisingly few works conducting visual analysis to explain why the learned similarity of a given image pair is high or low. \textit{Specifically, which part contributes the most to the similarity is a straightforward question and the answer can reveal important hidden information about the model as well as the data}.

Previous visual explanation works mainly focus on the interpretation of deep neural network for classification \cite{guided_bp,CAM, dissection, perturbation}. Guided back propagation (guided BP) \cite{guided_bp} has been used for explanation by generating the gradient from prediction to input, which shows how much the output will change with a little change in each dimension of the input. Another representative visual explanation approach, class activation map (CAM) \cite{CAM}, generates the heatmap of discriminative regions corresponding to a specific class based on the linearity of global average pooling (GAP) and fully connected (FC) layer. However, the original method only works on this specific architecture configuration (i.e. GAP+FC) and needs re-training for visualizing other applications. Based on the gradient of the last convolutional layer instead of the input, Grad-CAM \cite{Grad-CAM} is proposed to generate activation maps for all convolutional neural network (CNN) architectures. Besides, other existing methods explore network ablation \cite{dissection}, the winner-take-all strategy \cite{c-MWP}, inversion \cite{inverting}, and perturbation \cite{perturbation} for visual explanation.

Since verification applications like person re-identification (re-id) \cite{Re-ID} usually train metric learning models along with classification, recent work \cite{re_id_cam} leverages the classification activation map to improve the performance, but the activation map of metric learning is still not well explored. For two given images, a variant of Grad-CAM has been used for visualization of image retrieval \cite{re_id_grad} by computing the gradient from the cosine similarity of the embedding features to the last convolutional layers of both images. However, Grad-CAM only provides the overall highlighted regions of two input images, the relationship between each activated region of two images is yet to be uncovered. Since the similarity is calculated from two images and possibly based on several similar patterns between them, the relationship between these patterns is critical for understanding the model.

\begin{figure}
\begin{center}
\includegraphics[width=1.0\linewidth]{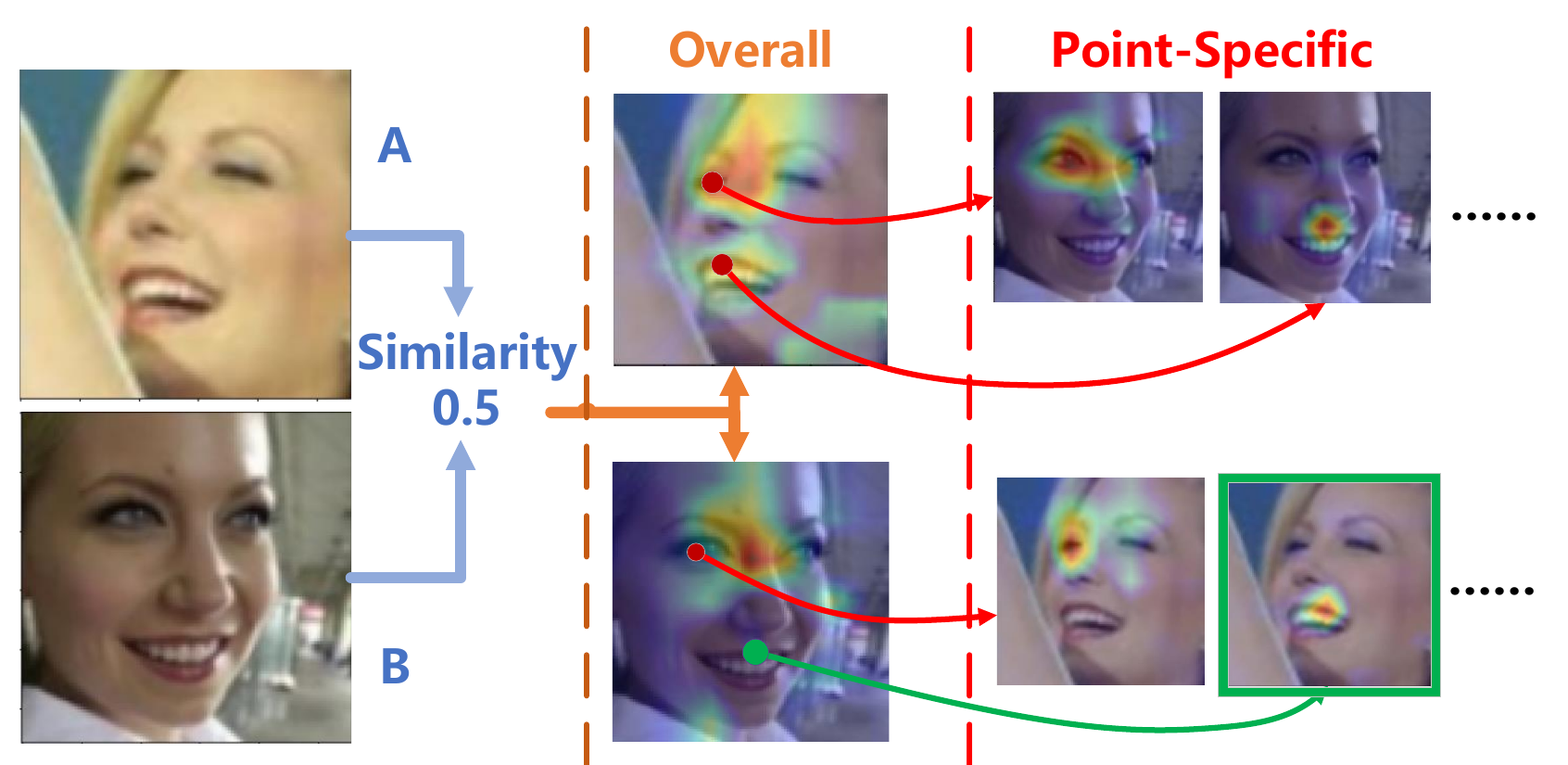}
\end{center}
\caption{An overview of activation decomposition. The overall activation map (the second column) on each image highlights the regions contributing the most to the similarity. The point-specific activation map (the third column) highlights the regions in one image that have large activation responses on a specific position, e.g. mouth or eye, in the other image.}
\label{fig:1}
\end{figure}

In this paper, we propose an \textbf{activation decomposition} framework for visual explanation of deep metric learning and explore the relationship between each activated region by \textit{point-to-point activation} response between two images. As shown in Fig. \ref{fig:1}, the overall activation maps (the second column) of two input images are generated by decomposing their similarity (score) along each image. In this example, the query image (A) has a high/strong activation on \textbf{both the eyes and mouth areas}, but the overall map of the retrieved image (B) \textbf{only highlights the eyes}. Therefore, it is actually hard to understand how the model works only based on the overall maps. 
For image B, the mouth region (\textcolor{ao}{green point}) has a low activation which means the activation between the mouth and \textbf{the whole image A} is low compared to the overall activation (similarity). 
However, by further decomposing this activation (\textcolor{ao}{green point}) along image A, that is to compute the activation between the mouth region of image B and \textbf{each position in image A}, the resulting \textit{point-specific} activation map (\textcolor{ao}{green box}) reveals that the mouth region of image B still has a high response on the mouth region of image A. This point-specific activation map can be generated for each pixel, which encodes the point-to-point activation intensity for representing the relationship between regions in both images, e.g. eye-to-nose or mouth-to-mouth. Compared with the overall activation map, the point-specific activation map provides much more refined information about the model which is crucial for explanation. 

The main contributions of this paper are summarized as follows.
\begin{itemize}[leftmargin=*]
   \item We propose a novel explanation framework for deep metric learning architectures based on activation decomposition. It can be served as a white-box explanation tool to better understand the metric learning models without modifying the model architectures, and is applicable to
   a host of metric learning applications, e.g. face recognition, person re-identification, image retrieval, etc.
   \item The proposed point-specific activation map uncovers the point-to-point activation intensity which, to the best of our knowledge, has not been explored by existing methods. Our experiments further show the importance of the point-specific activation map on two practical applications, i.e. cross-view pattern discovery and interactive retrieval. 
   \item We provide both theoretical and empirical analysis to show the superiority of the proposed overall activation map for deep metric learning over the popular Grad-CAM algorithm.

\end{itemize}

The remainder of this paper is organized as follows. Section \ref{sec:related} provides a brief review on existing visual interpretation methods for classification and metric learning. Section \ref{sec:naive} introduces the idea of activation decomposition for visual interpretation on a simple network architecture. 
Section \ref{sec:extension} further extends the framework to more complex architectures for metric learning as a unified solution. Section \ref{sec:gradcam} presents a detailed comparison between the proposed method and Grad-CAM. 
Section \ref{sec:experiment} validates the effectiveness of the overall activation map of our method, and Section \ref{sec:eva_point} demonstrates the unique advantages of the point-specific activation map on two practical applications.
Finally, Section \ref{sec:conclusion} concludes the paper.

\section{Related Work}
\label{sec:related}

\subsection{Interpretation for Classification}
\label{sec:classification}
As a default setting, most existing interpretation methods \cite{CAM, guided_bp, c-MWP, perturbation} are designed in the classification context where the output score is generated from one input image. Among the approaches that do not require architecture change, one intuitive idea \cite{CAM, c-MWP} is to look into the model and see where contributes the most to the final prediction. CAM \cite{CAM} and its variant Grad-CAM \cite{Grad-CAM} are among the most widely used approaches, but recent works \cite{perturbation,Grad-CAM,zhou2018interpretable} simply consider CAM as a heuristic linear combination of convolutional feature maps which is limited to the original architecture, global average pooling (GAP) and one FC layer \cite{CAM, Grad-CAM, zhou2018interpretable}. Specifically, to generate CAM for standard VGG \cite{VGG} with global max pooling (GMP) and three FC layers, \cite{CAM,zhou2018interpretable} have to replace this architecture with GAP+FC and re-train the model.
In this paper, we show that CAM is actually a \textit{special case} of activation decomposition on the specific architecture, and the activation decomposition idea applies to more complex architectures without re-training, even beyond the classification problem.

Grad-CAM \cite{Grad-CAM} is considered as a generalization of CAM for arbitrary CNN architectures, but the original paper only provides the proof on CAM's architecture. We show that Grad-CAM is not equivalent to activation decomposition for some architectures (Section \ref{sec:gradcam}). 

Another direction of explanation is to check the black box model by modifying the input and observing the response in the prediction. A representative method \cite{perturbation} aims to optimize the blurred region and see which region has the strongest response on the output signal when it is blurred. Perturbation optimization is applicable for any black box model, but the optimization is computationally expensive.

\begin{figure*}[htbp]
\begin{center}
\includegraphics[width=0.9\linewidth]{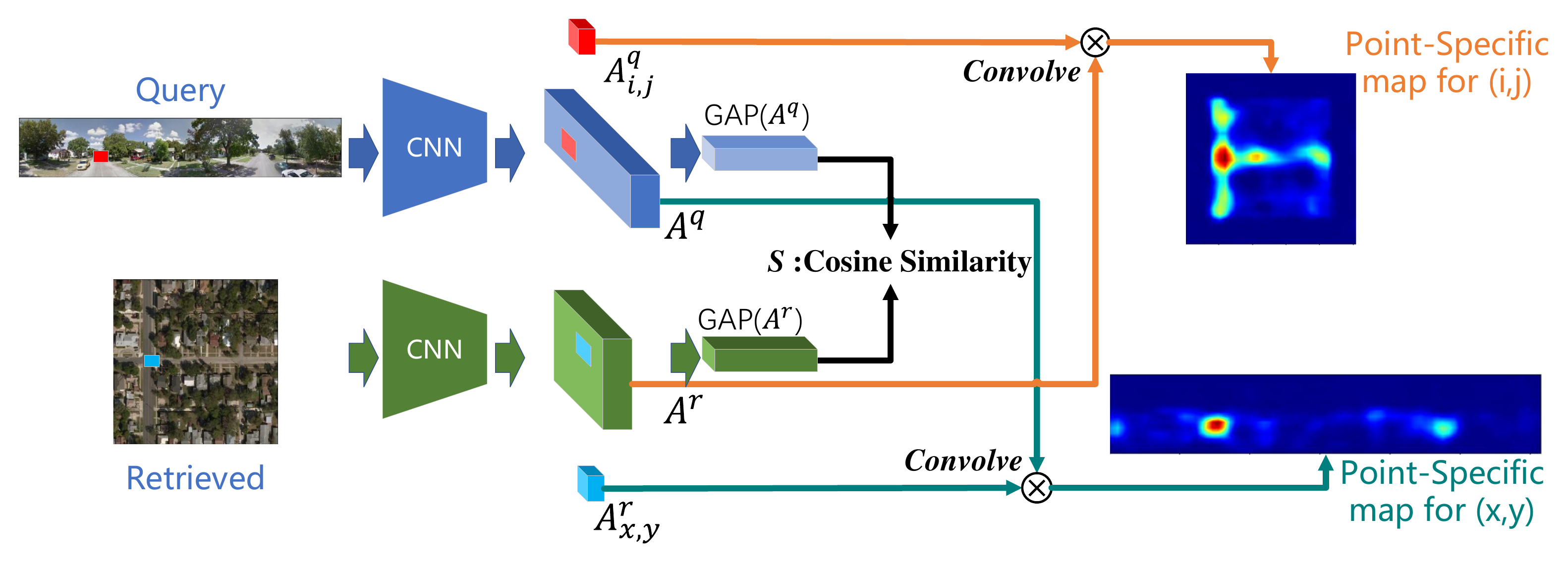}
\end{center}
\caption{Activation decomposition on a simple architecture for deep metric learning.}
\label{fig:2}
\end{figure*}

\subsection{Interpretation for Metric Learning}
As ignored by most existing methods, the explanation for metric learning is important for a wide range of applications \cite{wacv,chen2020adapting}, e.g. weakly supervised localization. In this work, we further shows its potential significance on diagnosis of metric learning losses and new applications including cross-view pattern discovery and interactive retrieval for person re-identification \cite{ye2020augmentation,ye2020probabilistic,ye2021deep,ye2020cross,ye2020visible} and face recognition. Gradient-based methods, e.g. guided BP \cite{guided_bp} can be adopted to metric learning, but recent works \cite{check,perturbation} claim that gradient can be irrelevant to the model and guided-BP fails on the sanity check \cite{check}. Among methods that pass the sanity check, Grad-CAM has been used for visualization of image retrieval in \cite{re_id_grad}, but no quantitative result is provided in the paper. Instead of adding more gradient heuristics like Grad-CAM++ \cite{Grad-CAM++}, we propose to explain the result of Grad-CAM under the decomposition framework. As shown in Section \ref{sec:gradcam}, by removing the heuristic step in Grad-CAM, the meaning of the generated activation map can be clearly explained using the proposed framework. 

Recent work \cite{wacv} also explores visualization techniques on metric learning, but it only provides qualitative results on several specific architectures. Another variant \cite{chen2020adapting} replaces the classification score with triplet loss in Grad-CAM framework and deploys weights transfer from the training data in the inference phase. The method requires three samples to form the triplet and its usage in the inference phase is dependent upon the specifically designed transfer procedure, thus is not a general method given only the pretrained model. Moreover, all the aforementioned methods only focus on the overall activation map which is not satisfactory for understanding the model as discussed in Section \ref{sec:intro}. Apart from white-box explanation, black-box optimization method, such as perturbation\cite{perturbation}, needs reformulation for metric learning and can be more computationally expensive if the point-to-point activation map is desired. 

\section{Activation Decomposition on A Simple Architecture}
\label{sec:naive}

\revise{To better introduce our method, we first review the formulation of CAM and illustrate our idea based on a simple architecture (CNN+GAP in Fig. \ref{fig:2}) for metric learning.} \textit{Note that the proposed method is not limited to this architecture, the general formulation will be presented in Section \ref{sec:extension}}. As shown in Eq. \ref{eq:1}, CAM is actually a spatial decomposition of the prediction score of each class, and the original method only applies for CNN with GAP and one FC layer without bias. The decomposition clearly shows how much each part of the input image contributes to the overall prediction of one class and provides valuable information about how the decision is made inside the classification model. The idea is based on the linearity of GAP:
\begin{equation}
    \label{eq:1}
    \small
    S_{c} = \sum_{k}\omega_{k,c}\left(\frac{1}{Z}\sum_{i,j}A_{i,j,k}\right) = \frac{1}{Z}\sum_{i,j}\left(\sum_{k}\omega_{k,c}A_{i,j,k}\right).
\end{equation}
$S_{c}$ denotes the overall score (before softmax) of class $c$ and $\omega_{k,c}$ is the FC layer parameter for the $k$-th channel of class $c$. $A_{i,j,k}$ denotes the feature map of the last convolutional layer at position $(i,j)$, and $Z$ is the normalization term of GAP. \reviseminor{Here, $Z$ equals to $m\times n$, which is the spatial size of feature map $A\in\R^{m\times n \times p}$}. In fact, the result $\sum_{k}\omega_{k,c}A_{i,j,k}$ can be considered as a decomposition of $S_{c}$ along $(i,j)$. For a two-stream Siamese-style architecture (see Fig. \ref{fig:2}) with GAP and the cosine similarity metric ($S$) for metric learning applications (e.g. image retrieval), we propose to do decomposition along $(i,j,x,y)$ so that the relationship between different parts of two images are uncovered:
\begin{equation}
\label{eq:2}
\begin{split}
    S & =  \reviseminor{(E^{q} \cdot E^{r})/|E^{q}||E^{r}|} \\
     & =  \sum_{k}GAP(A^{q}_{k})GAP(A^{r}_{k}) \reviseminor{/|E^{q}||E^{r}|} \\
    & =\frac{1}{Z}\sum_{k}\left(\sum_{i,j}A^{q}_{i,j,k}\sum_{x,y}A^{r}_{x,y,k}\right) \\ 
    & =\frac{1}{Z}\sum_{i,j,x,y}\left(\sum_{k}A^{q}_{i,j,k}A^{r}_{x,y,k}\right).
\end{split}
\end{equation}
\reviseminor{Here, $E^{q}$,$E^{r}$ denote the embedding feature of query and reference image. $|E|$ denotes L2 norm of $E$ and $\cdot$ is the inner product. The normalization term $Z$ equals to $m^{q}n^{q}m^{r}n^{r}|E^{q}||E^{r}|$, if the spatial size of $A^{q},A^{r}$ are $m^{q}\times n^{q}, m^{r}\times n^{r}$.} We use $(i,j)$ and $(x,y)$ for different streams because some applications such as cross-view image retrieval \cite{cvmnet} may have different image sizes for two streams. $A$ denotes the feature map of the last convolutional layer. The superscripts $q$ and $r$ respectively denote the query and retrieved images in this paper. For each query point $(i,j)$ in the query image, the corresponding \textbf{point-specific activation map} in the retrieved image is given by $\sum_{k}A^{q}_{i,j,k}A^{r}_{x,y,k}$ which is the contribution of features at $(i,j,x,y)$ to the overall cosine similarity. \revise{Like CAM and Grad-CAM, bilinear interpolation is implemented to generate the contribution of each pixel pair and the full resolution map.} The \textbf{overall activation maps} of two images are generated by a simple summation along $(i,j)$ or $(x,y)$, e.g. $\sum_{i,j}(\sum_{k}A^{q}_{i,j,k}A^{r}_{x,y,k})$. Although we only show the positive activation in the map, the negative value is also available for counterfactual explanation like \cite{Grad-CAM}.

Recent works \cite{sphere, cosine} have highlighted the superiority of L2 normalization, we therefore adopt the cosine similarity $S$ as the default metric. With L2 normalization, the squared Euclidean distance $D$ equals to $2-2S$ so that $S$ and $D$ are equivalent as metrics. Although there are still a number of methods \cite{Re-ID,re_id_cross} utilizing the Euclidean distance without L2 normalization, \cite{Re-ID} shows that the cosine similarity performs well as the evaluation metric. We further empirically show that the cosine similarity works well for explanation in this case (Section \ref{sec:interactive}).

\section{Generalization for Complex Architectures}
\label{sec:extension}
\revise{Recent metric learning approaches usually leverage more complex architectures to improve performance, e.g. adding several FC layers after the flattened feature or global pooling. Although different metric learning applications have different head architectures (e.g. GAP+FC) on CNN, the basic components are highly similar. 
Given the introduction of activation decomposition on a simple architecture in the previous section, we present an unified extension to make our method applicable to most existing state-of-the-art CNN architectures for various applications, including image retrieval \cite{MS}, face recognition \cite{Arcface}, person re-identification \cite{Re-ID}, and image geo-localization \cite{cvmnet,zhu2020vigor}.} 
Specifically, in Section \ref{sec:linear}, we address linear components by considering them together as one linear transformation. Then Section \ref{sec:nonlinear} focuses on transforming nonlinear components to linear ones in the \textit{inference phase}.

\subsection{Linear Component}
\label{sec:linear}

For the last convolutional layer feature $A\in \R^{m\times n\times p}$ ($m$, $n$ and $p$ denote the width, height, and number of channels), its GAP (global average pooling) is equivalent to the flattened feature $\hat{A}\in \R^{mnp}$ multiplied by a transformation matrix $T_{GAP} \in \R^{p \times mnp}$:  
\begin{equation}\label{eq:0-1}
    GAP(A) = \frac{1}{mn}\sum_{i,j}A_{i,j} =  T_{GAP}\hat{A}.
\end{equation}
Here $(i,j)$ denotes the spatial coordinates. By reshaping $T_{GAP}$ to $p\times m \times n\times p$ as $T^{*}_{GAP}$, the matrix is given by:
\begin{equation}\label{eq:0-2}
T_{GAP}^{*}(k',i,j,k) = \left\{
\begin{aligned}
\frac{1}{mn} & & k'=k \\
0 & & k'\neq k 
\end{aligned}
\right.
\end{equation}
The matrix $T_{GAP}$ is simply calculated by reshaping $T^{*}_{GAP}$ to $p\times mnp$. 
Therefore, the GAP can be considered as a special case of flattened layer multiplied by a transformation matrix.
Without loss of generality, we consider a two-stream framework with flattened layer followed by one FC layer with weights $\hat{W}\in\R^{l \times mnp}$ and biases $B\in \R^{l}$ ($l$ denotes the length of the feature embedding vector). Since the visual explanation is generated in the inference phase when typical components such as FC layer and batch normalization (BN) are linear, all the linear components together are formulated as one linear transformation $g(\hat{A}) = \hat{W}\hat{A} + B = \sum_{i,j} W_{i,j}A_{i,j}+B$ in the FC layer. Here $W_{i,j}\in\R^{l\times p}$ and $A_{i,j}\in \R^{p}$ denote the weights matrix and feature vector corresponding to position $(i,j)$. Although $B$ is ignored in CAM, we keep it as a residual term in the decomposition. Then Eq. \ref{eq:2} is re-formulated as:
\begin{equation}
\label{eq:3}
\small
    \begin{split}
    SZ 
    & =g^{q}(A^{q})\cdot g^{r}(A^{r}) \\
    & = \left(\sum_{i,j}W^{q}_{i,j}A^{q}_{i,j}+ B^{q}\right)\cdot \left(\sum_{x,y}W_{x,y}^{r}A^{r}_{x,y}+B^{r}\right)\\
    &= \sum_{i,j,x,y}\underbrace{(W^{q}_{i,j}A^{q}_{i,j})\cdot (W^{r}_{x,y}A_{x,y}^{r})}_{\textbf{point-to-point activation}} +  \sum_{i,j}(W^{q}_{i,j}A^{q}_{i,j}\cdot B^{r}) \\
    & \quad + \sum_{x,y}(W^{r}_{x,y}A^{r}_{x,y}\cdot B^{q})+B^{q}\cdot B^{r}.
    \end{split}
\end{equation}
Here $Z$ denotes the normalization term for cosine similarity (L2 norm \reviseminor{of embedding features}), and $\cdot$ is the inner product. As can be seen from Eq. \ref{eq:3}, the decomposition of $S$ has four terms, and the last three terms contain the bias $B$. The first term clearly shows the activation response for location pair $(i,j, x,y)$ and the \textbf{point-specific} map for query position $(i,j)$ is given by $I(x,y) = (W^{q}_{i,j}A^{q}_{i,j})\cdot (W^{r}_{x,y}A_{x,y}^{r})$. The second and third terms correspond to the activation between one image and the bias of the other image. Although they can be considered as negligible bias terms, they actually contribute to the overall activation map. For the overall activation map of the query image, the second term can be included, because $W^{q}_{i,j}A^{q}_{i,j}\cdot B^{r}$ varies at different $(i,j)$ positions while the third and last terms stay unchanged. Similarly, the first and third terms are considered when calculating the overall map for the retrieved image. We investigate both settings about the bias term (i.e. w or w/o) on the overall activation map and they are referred to as ``Decomposition": $(W^{q}_{i,j}A^{q}_{i,j})\cdot (\sum_{x,y}W^{r}_{x,y}A^{r}_{x,y})$, and ``Decomposition+Bias" : $(W^{q}_{i,j}A^{q}_{i,j})\cdot (\sum_{x,y}W^{r}_{x,y}A^{r}_{x,y}+B^{r})$ in Section \ref{sec:experiment}. The last term is pure bias which is the same for every input image. Therefore, we ignore this term if not mentioned. 


\subsection{Non-linear Component}
\label{sec:nonlinear}
Non-linear transformation increases the difficulty of activation decomposition, our idea is to 
find its approximate linear transformation which can be directly integrated into the formulation of Section \ref{sec:linear}. For a non-linear transformation $f(x)$ with its current input data $x \in [x_{0}-\delta, x_{0}+\delta]$ ($\delta$ is a small number), the first two terms of the Taylor expansion of $f(x)$ can be considered as its linear transformation \textit{in the inference phase}:
\begin{equation}
\small
    f(x) = f(x_{0}) + \frac{f'(x_{0})}{1!}(x-x_{0})+... \\
    \approx f(x_{0}) + f'(x_{0})(x-x_{0}).
\end{equation}
Therefore, our method can generalize to more complex architectures 
as long as the network is differentiable, since the formulation of gradient $f'(x_{0})$ can be computed by chain rule.

\begin{figure}
\centering
\includegraphics[width=0.95\linewidth]{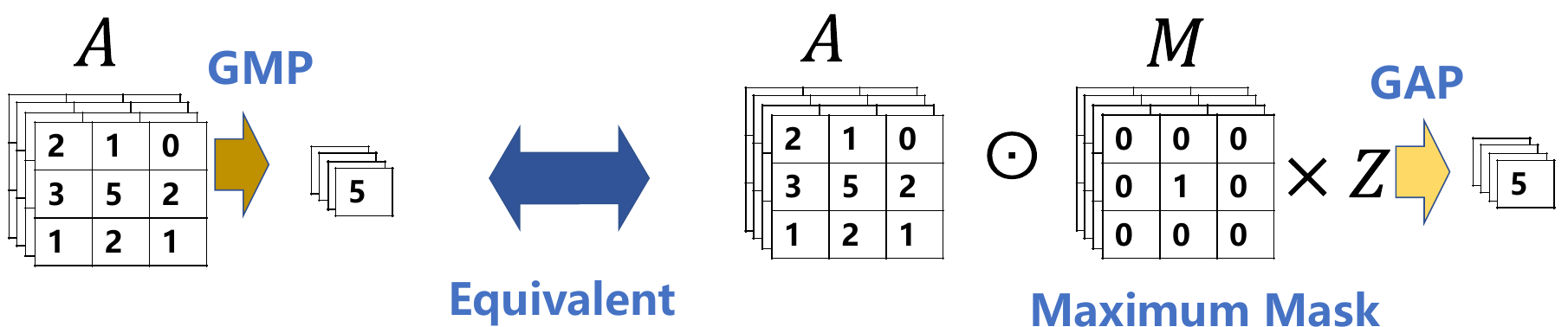}
\caption{Maximum mask for GMP. $Z$ is the constant normalization term for GAP.}
\label{fig:3}
\end{figure}

In this paper, we shed light on the practical guidelines for deploying our method on image retrieval, face recognition, person re-identification, and cross-view geo-localization frameworks. For these popular applications, the most widely used non-linear components -- global maximum pooling (GMP) and rectified linear unit (ReLU) -- can be transformed as linear operations {in the inference phase} by multiplying a mask\footnote{The mask is computed once for each input image.}.
In the inference phase, GMP can be considered as a combination of a maximum mask $M$ and GAP as shown in Fig. \ref{fig:3}. 
The transformation matrix of GMP is given by
\begin{equation}
\label{eq:GMP_trm}
    T^{*}_{GMP} = mn(T_{GAP}^{*}\odot M).
\end{equation}
$M\in R^{m\times n \times p}$ is the maximum matrix of $A$ where only the maximum position of each channel has nonzero value as $1$. $T_{GMP}$ is computed by reshaping $T^{*}_{GMP}$ to $p\times mnp$.
The result of Hadamard product ($M\odot A$) is considered as the new feature map which can be directly applied to Eq.~\ref{eq:3}.
Similarly, by adding a mask for ReLU, the FC layer with ReLU can be included in $W$ and $B$ in the inference phase. 

\section{Comparison with Grad-CAM}
\label{sec:gradcam}
\revise{In this section, we provide a comprehensive comparison between the proposed method and Grad-CAM with theoretical derivation.} Apparently, the proposed method can generate point-specific activation map which uncovers more fine-grained information than Grad-CAM, \revise{as Grad-CAM only generates  overall activation map.} We will further show the relationship between our overall activation map and Grad-CAM. As another way for generating overall activation map on more complex architectures, Grad-CAM follows a heuristic design without a clear meaning for the generated map. When we calculate the activation map based on Grad-CAM, what do we get? In \cite{Grad-CAM}, it has shown that Grad-CAM is equivalent to CAM on GAP based architecture, while it is not true on other architectures. For a classification architecture with flattened feature and one FC layer, the prediction score for class $c$ is formulated as $S_{c}=\sum_{i,j,k}W_{i,j,k,c}A_{i,j,k}$, where $W$ is the reshaped weights of the FC layer following the shape of the last convolutional feature ($A$). From our activation decomposition perspective, the contribution of each position $(i,j)$ is clearly given by $\sum_{k}W_{i,j,k,c}A_{i,j,k}$, while the Grad-CAM map for class $c$ at $(i,j)$ position is given by:
\begin{equation}
    \label{eq:4}
    \begin{split}
    GradCAM_{i,j,c} & = \sum_{k} A_{i,j,k}GAP\left(\frac{\partial S_{c}}{\partial A_{k}}\right)\\
    & =\sum_{k} A_{i,j,k}GAP\left(W_{k}\right).
    \end{split}
\end{equation}
In this case, the result of Grad-CAM is different from activation decomposition because of the GAP operation in Eq. \ref{eq:4}. For architectures using GMP, they are also different, because only the maximal value of each channel contributes to the overall prediction score with the activation decomposition framework, while Grad-CAM puts the same weight for features at all positions. 
Concretely, for a classification architecture with GMP and one FC layer ($W\in \R^{p\times l}$ and no bias), the prediction score for class $c$ is given by $S_{c}=\sum_{k} max_{i,j}(A_{:,:,k})W_{k,c}=\sum_{i,j,k}A_{i,j,k} M_{i,j,k}W_{k,c}$, where $M$ is the maximum matrix in Eq. \ref{eq:GMP_trm}. The decomposition of position $(i,j)$ is $\sum_{k}A_{i,j,k}M_{i,j,k}W_{k,c}$ and only the maximum position of each channel has a nonzero value. However, for Grad-CAM, the activation map is given by:
\begin{equation}
    GradCAM_{i,j,c}=GAP\left(\frac{\partial S_{c}}{\partial A}\right)A_{i,j} =\frac{1}{mn}\sum_{k}A_{i,j,k}W_{k,c},
\end{equation}
where all $(i,j)$ positions have a nonzero weight resulting in a more scattered activation map.

\begin{table*}[t!]
    \centering
    \caption{Weakly supervised localization accuracy (IOU=0.5) with different thresholds on CUB validation set.}
    \begin{tabular}{c|c|c|c|c}
    \hline
    Threshold & Grad-CAM     & Grad-CAM (no norm) &  Decomposition+Bias (ours) &  Decomposition (ours) \\
    \hline
    0.15 & \textbf{16.71}\% & 16.83\% & 23.49\% & 17.38\% \\
    0.2 & 16.26\% & 17.06\% & 32.01\% & 19.77\% \\
    0.3 & 14.48\% & 21.59\% & \textbf{44.94}\% & 34.92\% \\
    0.4 & 9.43\% & 35.01\% & 37.85\% & \textbf{50.64}\% \\
    0.5 & 4.43\% & 47.00\% & 21.99\% & 45.78\% \\
    0.6 & 1.54\% & \textbf{48.27}\% & 9.35\% & 27.98\% \\
    0.7 & 0.27\% & 20.90\% & 2.25\% & 9.11\% \\
    \hline
    \end{tabular}
    \label{table:loc_threshold}
\end{table*}

\begin{figure}[htbp]
\centering
\includegraphics[width=0.99\linewidth]{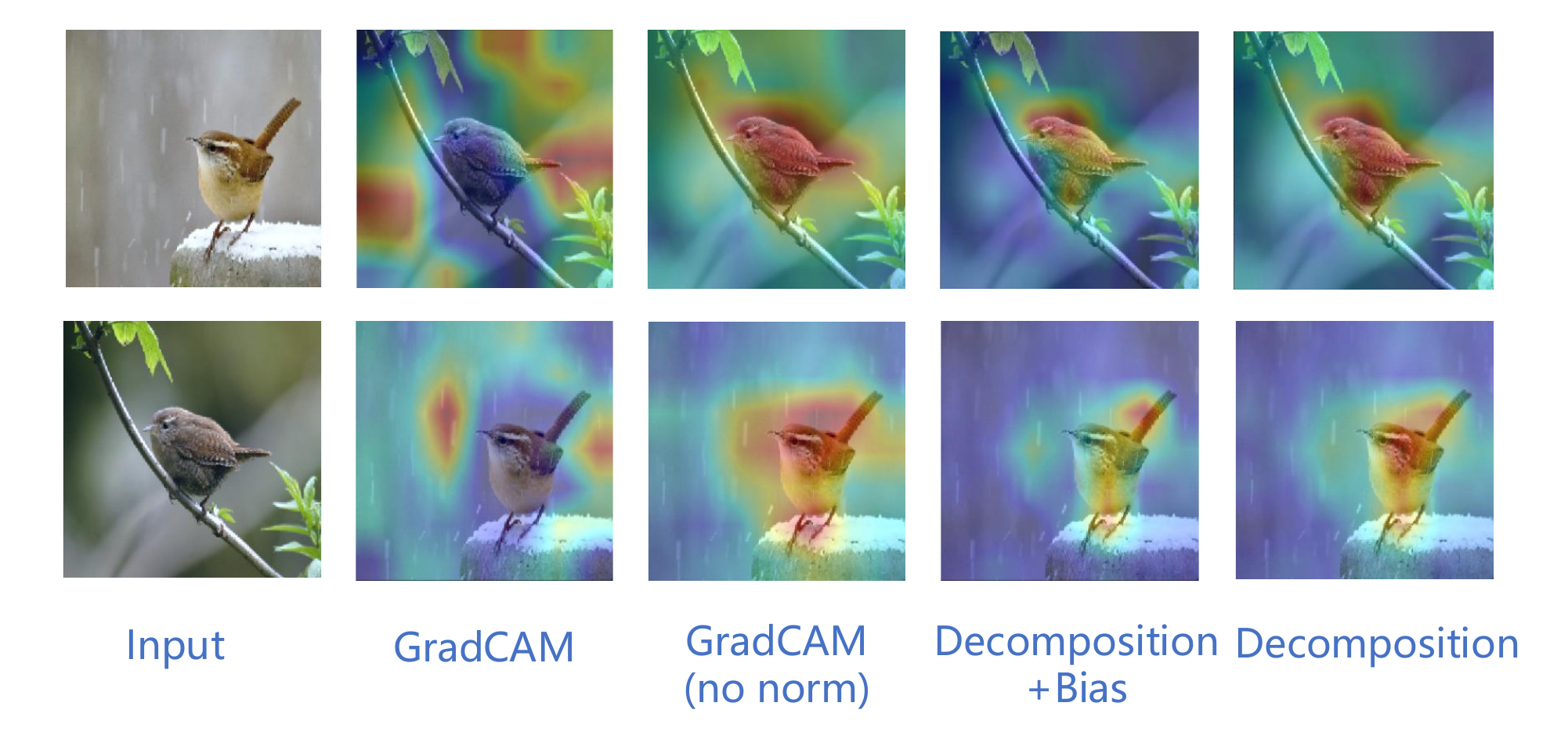}
\caption{The activation maps of different approaches on an example image pair from the CUB dataset.}
\label{fig:4}
\end{figure}

\revise{Although \cite{Grad-CAM} empirically shows that the heuristic GAP step can help improve the performance of object localization, this step makes the meaning of the generated map unclear.} By removing this step, which means combining the gradient and feature directly as $A_{i,j,k}\frac{\partial S_{c}}{\partial A_{i,j,k}}$, the modified version would generate the same result as activation decomposition for flattened and GMP based architectures. As for the metric learning architecture discussed in Section \ref{sec:linear}, the Grad-CAM map of query image is written as (\textit{the detailed derivation is presented in Appendix \ref{app:gradcam_metric}}):
\begin{equation}
    \label{eq:5}
    \begin{split}
    GradCAM_{i,j} =& 
    \frac{1}{Z}\left(\frac{\partial (E^{q}/|E^{q}|)}{\partial E^{q}}GAP(W^{q}) A^{q}_{i,j}\right) \\
    & \cdot \underbrace{\left(\sum_{x,y}W^{r}_{x,y}A^{r}_{x,y}+B^{r}\right)}_{E^{r}}.
    \end{split}
\end{equation}
$E^{q}$ and $E^{r}$ denote the embedding vectors of the query and retrieved images before L2 normalization and $Z$ is the normalization term. \revise{The gradient term $\frac{\partial (E^{q}/|E^{q}|)}{\partial E^{q}}$, which comes from the L2 normalization, would put less weights on the dominant channels so that the generated activation map becomes more scattered as shown in Fig. \ref{fig:4} (\textit{please also refer to Appendix \ref{app:gradcam_metric} for proof}).} It can be removed by calculating gradient from $E^{q}\cdot E^{r}$ without L2 normalization, denoted as ``Grad-CAM (no norm)". If we remove the gradient term as well as the GAP term, the result is actually equivalent to the overall map of our ``Decomposition+Bias" given by the first two terms of Eq. \ref{eq:3} as $(W^{q}_{i,j}A^{q}_{i,j})\cdot (\sum_{x,y}W^{r}_{x,y}A^{r}_{x,y}+B^{r})$.

\section{Evaluating Overall Activation Map}
\label{sec:experiment}
To validate the advantage of the proposed method, we provide results of human evaluation as well as several practical applications. We first verify the effectiveness of the proposed overall activation map with weakly supervised localization and human evaluation (Sections \ref{sec:weakly} and \ref{sec:human}). We also show how the activation map brings potential insights on the generalization ability of different metric learning models (Section \ref{sec:diagnosis}).


\begin{table}[htbp]
\centering
\caption{Weakly supervised localization accuracy (IoU=0.5) of various methods on CUB validation set.}
\begin{tabular}{l|l|c}
\hline
& Method & Accuracy\\
\hline
Classification & CAM \cite{CAM} & 41.0\% \\
\hline
\multirow{4}*{Metric Learning}  & Grad-CAM & 16.7\%\\
\cline{2-3}
 & Grad-CAM (no norm) & 48.3\% \\
 & Decomposition+Bias (ours) & 44.9\% \\
 & Decomposition (ours) & \textbf{50.6\%} \\
\hline
\end{tabular}
\label{table:1}
\end{table}

\begin{figure*}[htbp]
    \centering
    \includegraphics[width=0.95\linewidth]{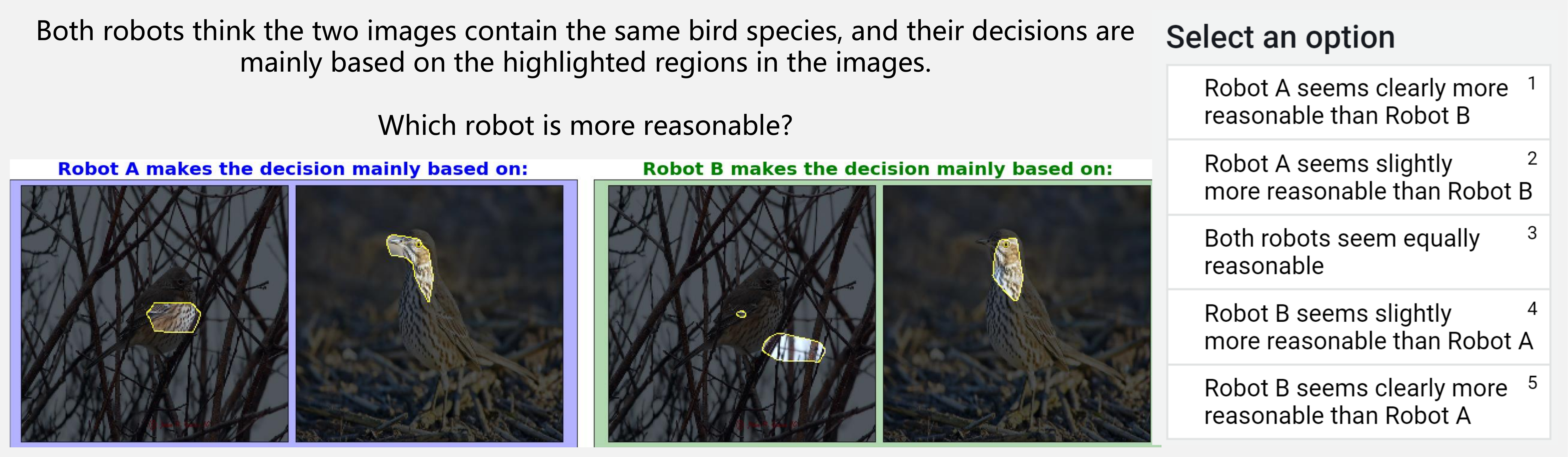}
    \caption{Human evaluation interface for evaluating the overall activation maps on CUB.}
    \label{fig:human_cub_interface}
\end{figure*}

\begin{figure*}[!htbp]
    \centering
    \includegraphics[width=0.9\linewidth]{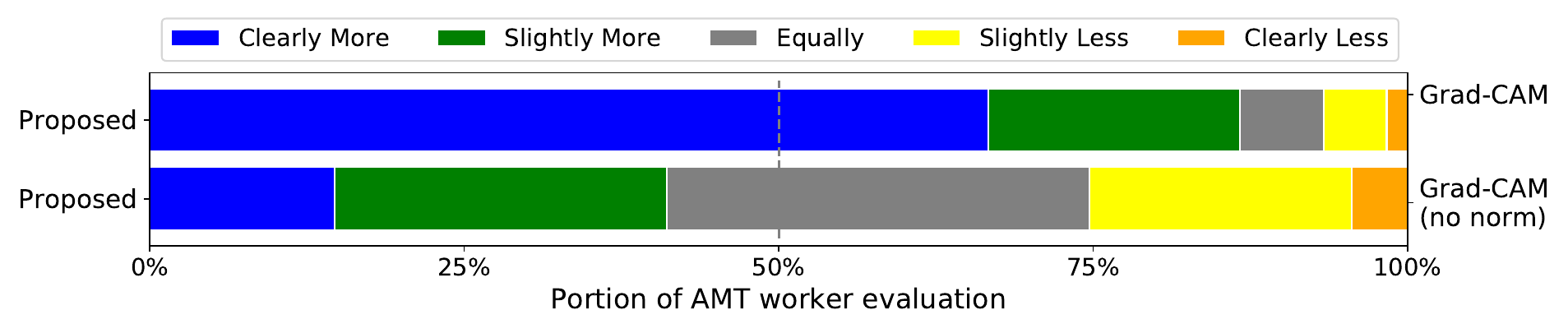}
    \caption{Human evaluation results on the proposed method vs. Grad-CAM variants. \reviseminor{The ``clearly more" corresponds to the portion of workers who select the option 1 in Fig. \ref{fig:human_cub_interface}, which means they think Robot A is clearly more reasonable than Robot B.}}
    \label{fig:human_cub}
    \vspace{-0.2cm}
\end{figure*}

\subsection{Weakly Supervised Localization}
\label{sec:weakly}

Following previous works \cite{CAM,Grad-CAM} on visual explanation, we conduct weakly supervised localization experiment in the context of metric learning. 
Among the popular datasets for image retrieval, CUB \cite{CUB} is a challenging dataset with over 10,000 images from 200 bird species, and the bounding box annotation is available for localization evaluation. We first train a metric learning model on CUB using the state-of-the-art image retrieval method \cite{MS}. Then the weakly supervised localization is conducted to show the differences between the variants of the proposed visual explanation method. 

We first generate the activation maps of different methods, then the mask is generated by segmenting the heatmap with a threshold. Following previous works \cite{CAM,Grad-CAM}, the bounding box that covers the largest connected component in the mask is generated as the prediction. A predicted bounding box is considered correct if the IoU (Intersection over Union) between the bounding box and ground-truth is greater than $0.5$. 
\revise{Apparently, different thresholds would lead to different predictions, we thus apply a grid search on the threshold to find the best performance of each competing method for a fair comparison.} Table \ref{table:loc_threshold} lists the localization accuracy on CUB validation set under different thresholds.
For convenience, we report the best result of each method in Table~\ref{table:1}, including the result of CAM (classification based method) adopted from the original paper \cite{CAM}.

In these two tables, ``Grad-CAM (no norm)" means Grad-CAM with gradient computed from the product of two embedding vectors without normalization and ``Decomposition+Bias" denotes the first two terms of Eq. \ref{eq:3}. Since the architecture of \cite{MS} is based on GMP, ``Grad-CAM (no norm)" is not equivalent to ``Decomposition+Bias" as shown in  Section \ref{sec:gradcam}. The proposed framework outperforms Grad-CAM for metric learning as well as CAM based on classification. As justified in Section \ref{sec:gradcam}, computing the gradients before normalization does improve the performance of Grad-CAM by a large margin. The qualitative result in Fig. \ref{fig:4} also illustrates the scattering effect of Grad-CAM with normalization as demonstrated in our theoretical analysis (Appendix \ref{app:gradcam_metric}). \revise{The information in the bias term does not help the weakly supervised localization, as witnessed by the performance of ``Decomposition+Bias".} 

\begin{figure*}[htbp]
\begin{center}
\includegraphics[width=1.0\linewidth]{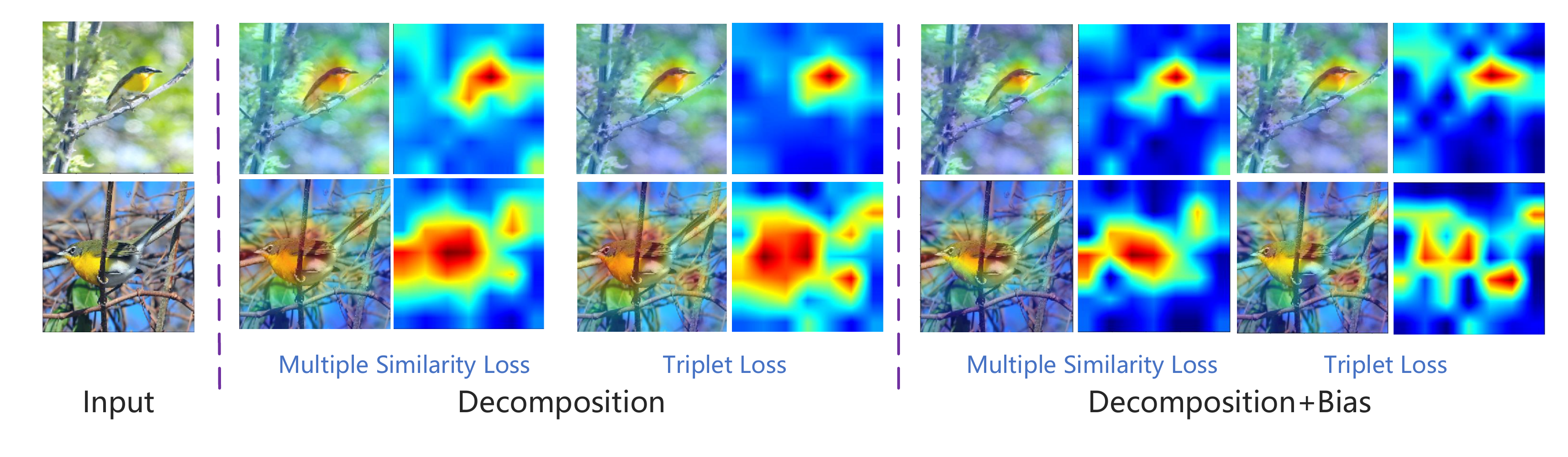}\vspace{-0.2cm}
\end{center}
\caption{Qualitative results (overall activation maps corresponding to different metric learning losses) for model diagnosis. Although the metric learning models with two different losses yield similar training accuracies (82.69\% and 82.98\% in Table \ref{table:2_1}), their overall activation maps generated by the activation decomposition framework reveal great differences in the regions to which the models pay attention. These activation maps also provide valuable insights on the model generalization ability to the validation/test set.}
\label{fig:5}
\end{figure*}

\subsection{Human Evaluation}
\label{sec:human}

Following the work of Grad-CAM \cite{Grad-CAM}, we also conduct human evaluation on the widely used platform (AMT: Amazon Mechanical Turk) to further evaluate the quality of overall activation maps generated by different methods. 
Based on Table \ref{table:1}, two comparison pairs are formed as ``Decomposition" vs. ``Grad-CAM" and ``Decomposition" vs. ``Grad-CAM (no norm)". For each comparison pair, the AMT raters are asked to carefully check the visual explanation generated from two different methods as shown in Fig. \ref{fig:human_cub_interface}. Each visual explanation contains an image pair with birds from the same species and the top activated regions ($2\%$ pixels) are highlighted based on the overall activation map of one competing method. We denote these two methods as robot A and robot B in the questionnaire. The raters will decide which robot they think is more reasonable in a five-point Likert scale as shown in Fig. \ref{fig:human_cub_interface}. \revise{The five scales denote method A is ``clearly more", ``slightly more", ``equally", ``slightly less", ``clearly less" reasonable than method B.}

$720$ evaluations of comparison pairs are collected from at least $40$ different AMT workers for two comparison pairs (half for each) and the result is consistent with the outcome of Section \ref{sec:weakly} as shown in Fig. \ref{fig:human_cub}. Over $60\%$ evaluations support that the proposed method is clearly more reasonable than the vanilla Grad-CAM. Removing the normalization for the gradient computing does improve the quality of Grad-CAM's activation map significantly, but the proposed method still holds more supportive evaluations ($41.1\%$ vs. $25.2\%$) in the AMT test.

\subsection{Model Diagnosis}
\label{sec:diagnosis}
A large number of loss functions have been proposed for metric learning \cite{FaceNet,Lifted,MS}, while only the performance and embedding distribution are evaluated. We show that the overall activation map can also help evaluate the generalization ability of different metric learning methods. 
\begin{table}[htbp]
\centering
\caption{Top-1 recall accuracy on CUB.}
\begin{tabular}{l|c|c}
\hline
Loss & Train & Validation\\
\hline
MS & 82.69\% & 65.45\% \\
Triplet & 82.98\% & 60.55\% \\
\hline
\end{tabular}
\label{table:2_1}
\end{table}

\begin{table}
\centering
\caption{Localization accuracy (IoU=0.5) on CUB training set.}
\begin{tabular}{l|c|c}
\hline
& Decomposition+Bias & Decomposition\\
\hline
MS & 52.10\% & 58.99\% \\
Triplet & 39.93\% & 55.60\% \\
\hline
Accuracy Drop & 12.17\% & 3.39\%\\
\hline
\end{tabular}
\label{table:2_2}
\end{table}

\begin{figure*}[htbp]
\centering
\includegraphics[width=1.0\linewidth]{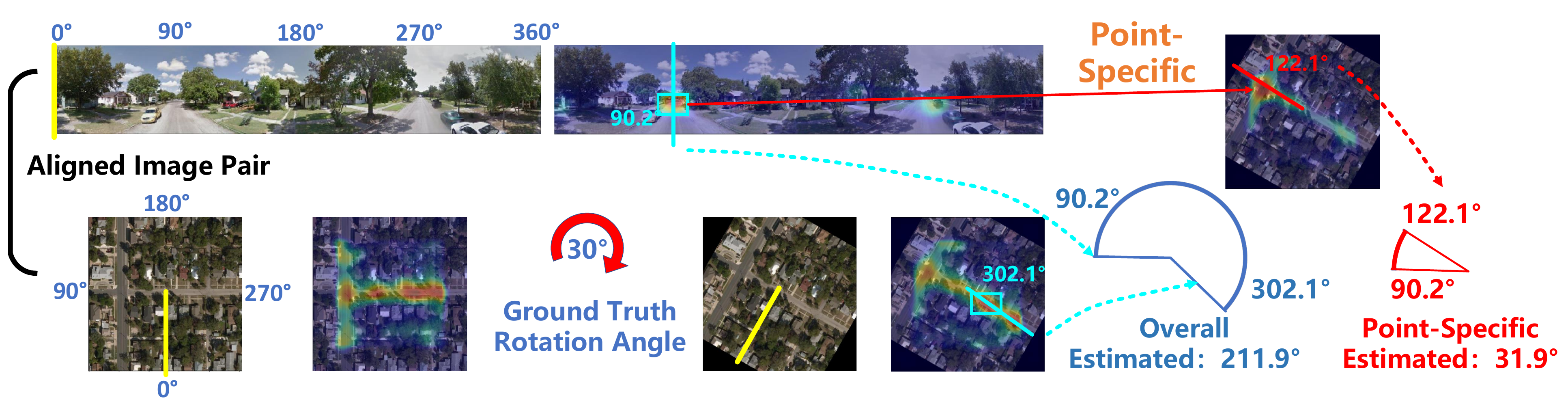}
\caption{An example of illustrating cross-view pattern discovery -- image orientation estimation -- by using the activation maps from the metric learning (i.e. image matching/retrieval) model. Based on the overall activation maps from two views, the estimated orientation angle between two views is $211.9^{\circ}$. By generating the point-specific activation map of a maximum activated region in the other view, the estimated orientation is much more accurate, i.e. $31.9^{\circ}$, with only $1.9^{\circ}$ of angle difference as compared with the ground truth in this example.} 
\label{fig:6}
\end{figure*}


We follow the setting of \cite{MS} and train two metric learning models with Multiple Similarity (MS) loss \cite{MS} and Triplet loss \cite{FaceNet}, respectively. As shown in Table \ref{table:2_1}, although they have almost the same accuracy on the training set, there is a big gap between their generalization abilities on the validation set. \textit{Before checking the result on the validation set, is there any clue in the training set for such gap?} Despite the similar training accuracies, the predictions of two models are actually based on different regions from the activation maps generated by our activation decomposition framework (see Fig. \ref{fig:5}). The activation maps also provide valuable insights on the model generalization ability to the validation/test set. It is evident from Fig. \ref{fig:5}, the model with MS loss is more focused on the salient object area than the one with Triplet loss, which leads to better validation performance as confirmed in Table \ref{table:2_1}. \revise{The quantitative results of localization in Table \ref{table:2_2} also support the fact that ``Triplet" model is more likely to focus on the background region rather than the object as we witness the localization accuracy drop for Triplet loss.} Although ``Decomposition+Bias" can be more sensitive to different loss functions, implying that the bias term does provide valuable information on the overall map, both settings of the proposed method have the same trend on accuracy. Therefore, our activation decomposition framework can shed light on the generalization ability of loss functions for metric learning, thereby providing an useful tool for model diagnosis.

\section{Evaluating Point-Specific Activation Map}
\label{sec:eva_point}

\begin{figure}
\centering
\includegraphics[width=0.95\linewidth]{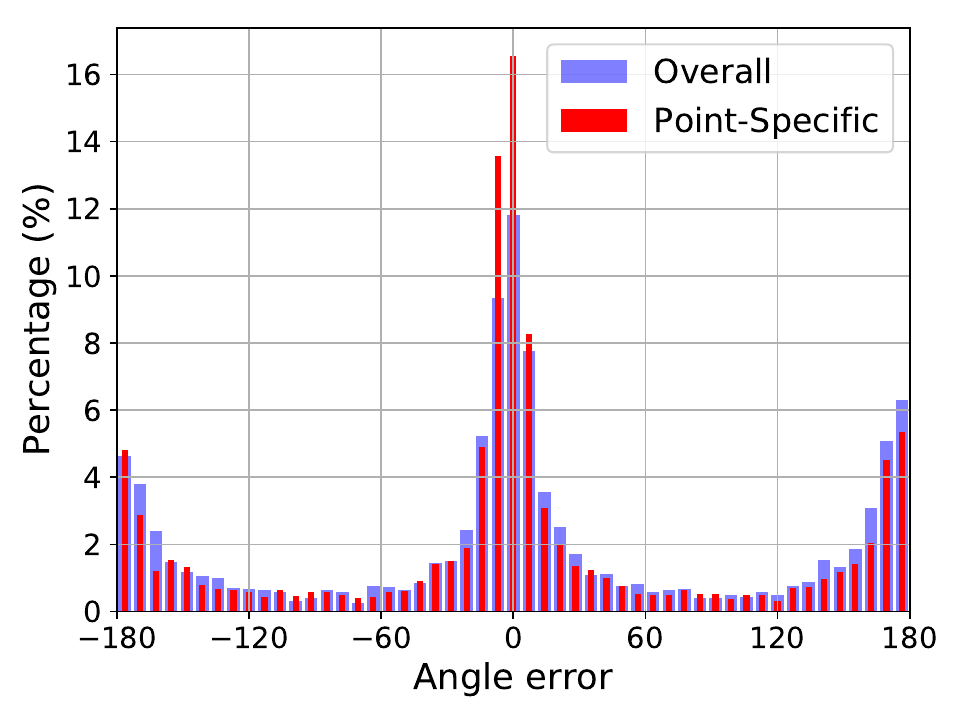}
\caption{The results of cross-view image orientation estimation (i.e. angle error distributions) obtained by the overall and point-specific maps based on the image matching experiment on the CVUSA dataset. }
\label{fig:7}
\end{figure}

The proposed activation decomposition framework not only generates the overall activation map, but also can produce the point-specific activation map for fine-grained visual explanation of metric learning. In this section, we introduce two applications to demonstrate the importance of the proposed point-specific activation map.

\subsection{Application I: Cross-view Pattern Discovery}
\label{sec:cross}

When metric learning is applied to cross-view applications, e.g. image retrieval \cite{CVUSA,cvmnet,chen}, the model is capable of learning similar patterns in different views which may provide geometric information of the two views, e.g. the camera pose or orientation information. We conduct orientation estimation experiment to show the advantage of point-specific activation map compared with the overall activation map on providing geometric information based on cross-view patterns. In our experiment, we take street-to-aerial image geo-localization \cite{CVUSA,cvmnet} as an example.

The objective of cross-view image geo-localization is to find the best matched \textit{aerial-view image} (with GPS information) in a reference dataset for a query \textit{street-view} image. We conduct the experiment on CVUSA \cite{CVUSA}, which is the most popular benchmark for this problem containing 35,532 training pairs and 8,884 test pairs. Each pair consists of a query street image and the corresponding \textbf{orientation-aligned} aerial image at the same GPS location. For example, as shown in Fig. \ref{fig:6}, the left two images (street-view panorama and aerial-view images) are the original aligned image pair and the yellow line denotes the $0^{\circ}$ which corresponds to the South direction ($180^{\circ}$ corresponds to the North direction). \revise{The location of panorama always lies at the center of the corresponding aerial image.} We use $[0,360]^{\circ}$ to denote different angles as marked on those two images.

\begin{figure*}[htbp]
\centering
\includegraphics[width=0.95\linewidth]{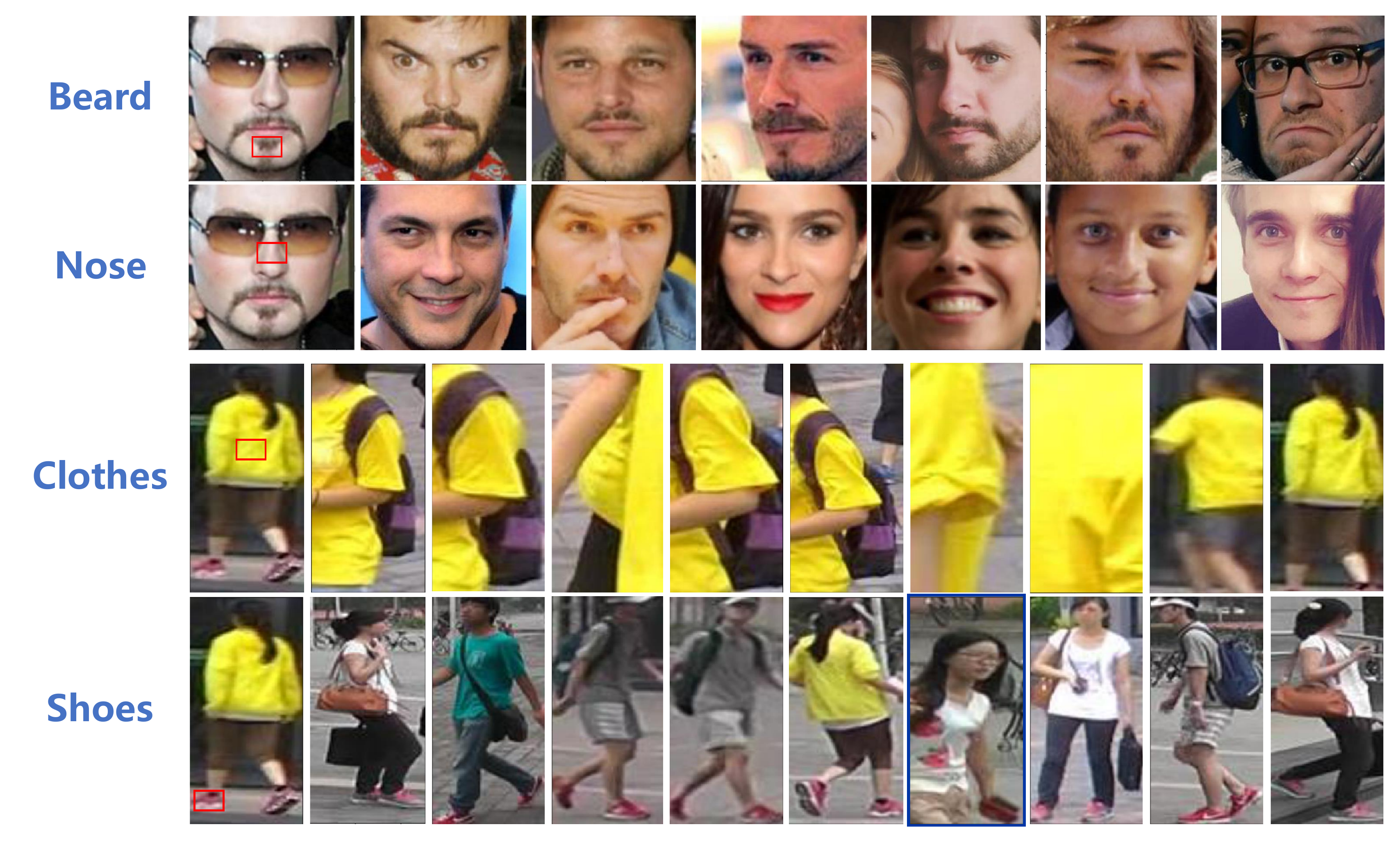}
\caption{Top retrieved images by interactive retrieval on face and re-id datasets. The first column shows the query images and the red box on each of them indicates the region of interest (RoI). In the last row, the blue bounding box highlights a failure retrieval case, which is further analyzed in Fig. \ref{fig:fail}.}
\label{fig:8}
\end{figure*}

We train a simple Siamese-VGG \cite{Zhu_2021_WACV,VGG} network following the modified triplet loss in \cite{cvmnet}. During the training, if the image pairs are not aligned (e.g. randomly rotate the aerial view images), the activation map can be used for orientation estimation. Specifically, we train the Siamese-VGG with randomly rotated aerial images so that the model is approximately rotation-invariant and so is the overall activation map for aerial image. In the example of Fig. \ref{fig:6}, the overall activation maps for the original aligned pair are first generated to show the highlighted regions contributing to the similarity score, and both views focus on the road areas in this example. The most activated regions are highly relevant in both views and most of them are similar patterns. When we randomly rotate the aerial image ($30^{\circ}$ in this example and is denoted as the ground truth rotation angle in Fig. \ref{fig:6}), the aerial view activation map still focuses on the road area which is relevant to the street view.

In this paper, we simply use the pixel with the maximum activation value for orientation estimation, because the most activated areas (highlighted in cyan boxes in Fig. \ref{fig:6}) from two views are likely to be relevant. For the street-view image, the selected pixel lies in the angle of $90.2^{\circ}$ (the cyan line) based on the angle marks on the left aligned image. For the rotated aerial view, the selected pixel lies in the angle of $302.1^{\circ}$ (the cyan line). Since the overall map contains multiple activated regions, the selected (maximum activation) pixels in both views actually do not correspond to the same object, which reveals one disadvantage of using only the overall activation map. In this example, the estimated angle is $302.1^{\circ}-90.2^{\circ} = 211.9^{\circ}$, which is not correct. However, for the selected pixel (the one with the maximum activation value) in the street-view image, if we generate its corresponding point-specific activation map on the aerial-view image (as shown on the very right of Fig. \ref{fig:6}), the new selected pixel (maximum activation) on this point-specific activation map lies in the angle of $122.1^{\circ}$ (the red line). Then, the estimated angle is calculated as $122.1^{\circ}-90.2^{\circ}=31.9^{\circ}$, which is very close to the ground truth ($30^{\circ}$). This demonstrates the advantage of the point-specific activation decomposition for finding more fine-grained information in this application. 

We also present the quantitative results in Fig. \ref{fig:7}. The angle error is computed by \texttt{error = ground truth - estimated angle}. We then add or subtract $360^{\circ}$ to the error to fit it in the range of $[-180,180]^{\circ}$ if the absolute value of the error is greater than $180^{\circ}$. For example, when the error is $359^{\circ}$, we subtract it by $360^{\circ}$ and get $-1^{\circ}$ as the error. This is a reasonable setting adopted from the previous work \cite{CVUSA}. As evident from Fig. \ref{fig:7}, the point-specific activation map significantly outperforms the overall map for cross-view orientation estimation. Point-specific activation based orientation estimation has over $16\%$ samples (a total of 8,884 test sample pairs in the CVUSA dataset) with angle error less than $\pm 3.5^{\circ}$ (the red bar at $0^{\circ}$ in Fig. \ref{fig:7}), while overall activation map based method has less than $12\%$. The result further validates the superiority of the point-specific map compared with the overall map.
Moreover, we also provide a demo\footnote{Please check the demo (the ``Geo-localization" section) at \url{https://github.com/Jeff-Zilence/Explain_Metric_Learning}} to show how the point-specific map changes according to the query pixel. \revise{Failure samples of both methods tend to have an angle error of 180 degree, because the activation maps of both views are likely to focus on roads which are symmetric in aerial images.}



\subsection{Application II: Interactive Retrieval}
\label{sec:interactive}

\begin{figure*}[htbp]
    \centering
    \includegraphics[width=0.95\linewidth]{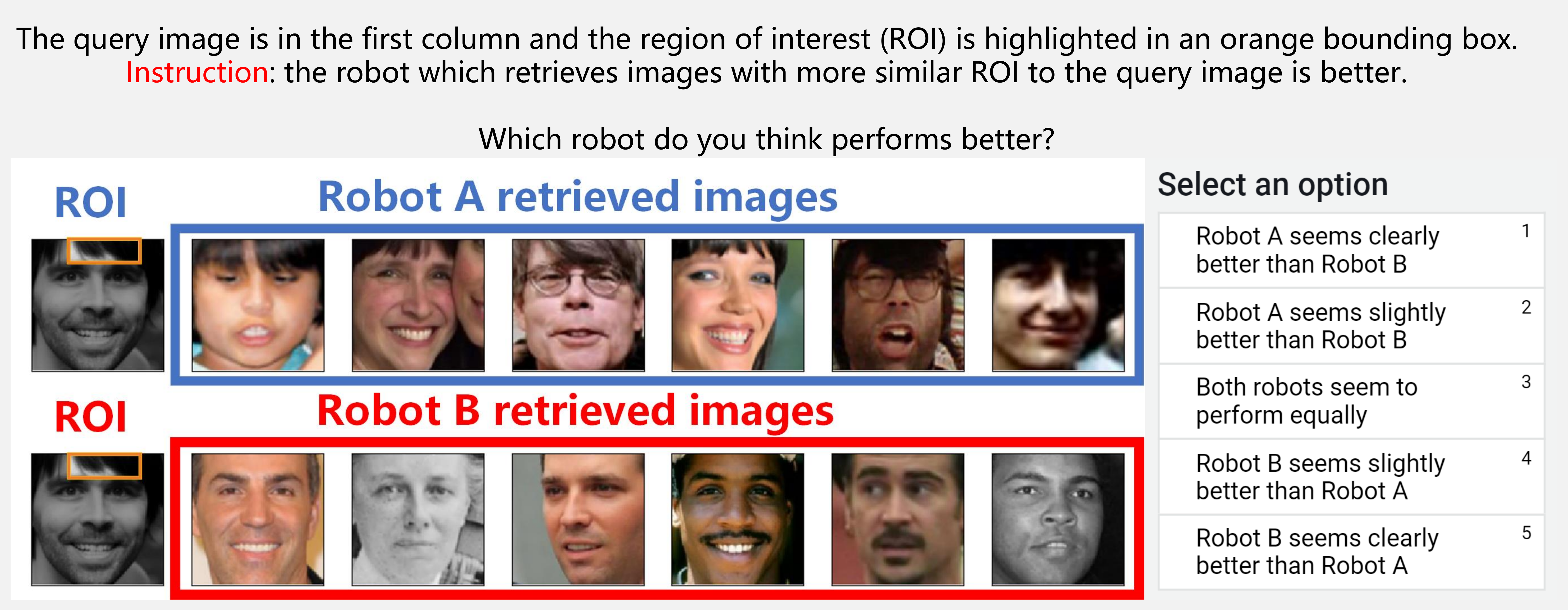}
    \caption{Human evaluation interface for interactive retrieval.}
    \label{fig:human_face_interface}
\end{figure*}

\begin{figure*}
  \begin{minipage}[b]{0.7\linewidth}
    \includegraphics[width=\textwidth]{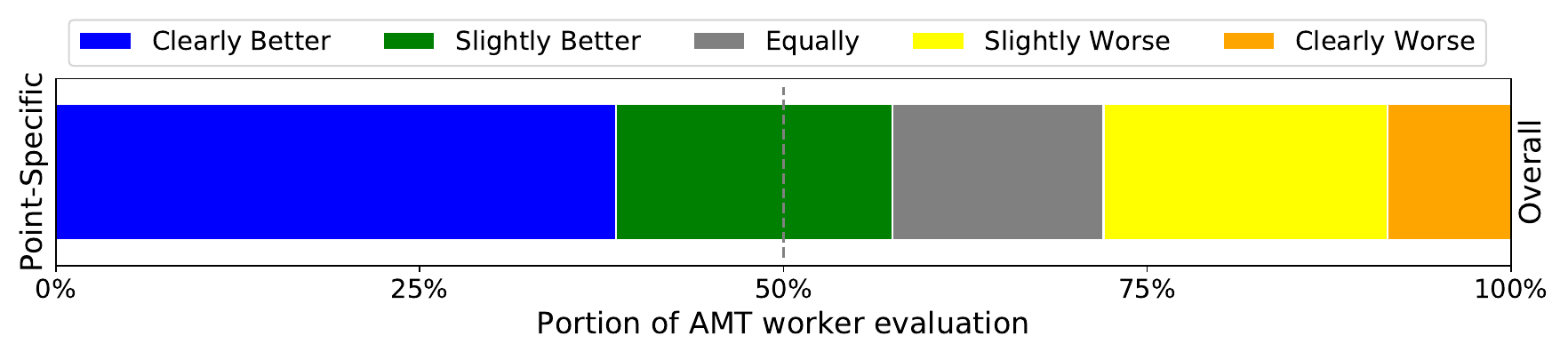}
    \includegraphics[width=\textwidth]{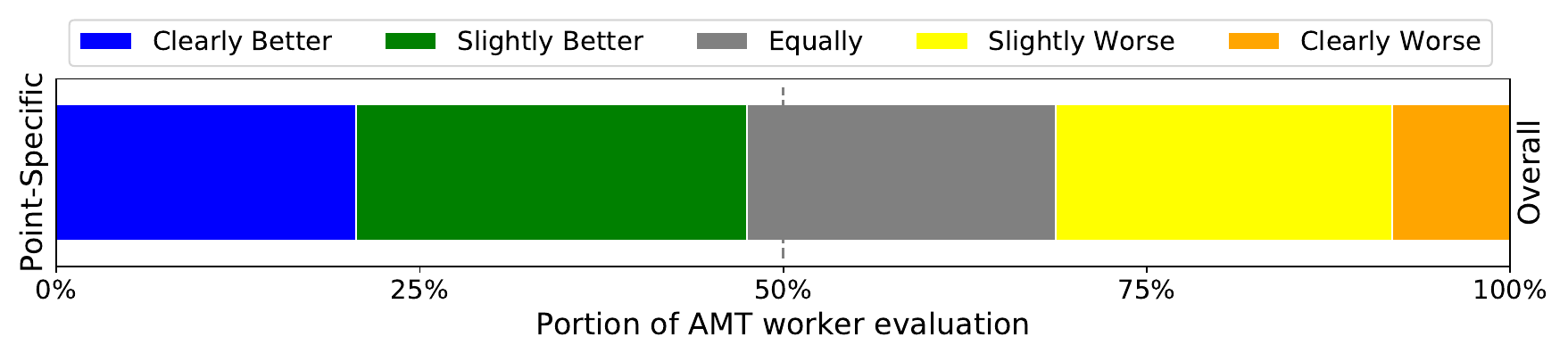}
    \caption{Human evaluation results on the point-specific (interactive) vs. overall retrieval for face verification (first row) and person re-identification (second row). \reviseminor{The ``clearly better" corresponds to the portion of workers who select the option 1 in Fig. \ref{fig:human_face_interface}, which means they think Robot A performs clearly better than Robot B.}}
    \label{fig:human_face_reid}
  \end{minipage}
  \quad
  \begin{minipage}[b]{0.3\linewidth}
    \includegraphics[width=\textwidth]{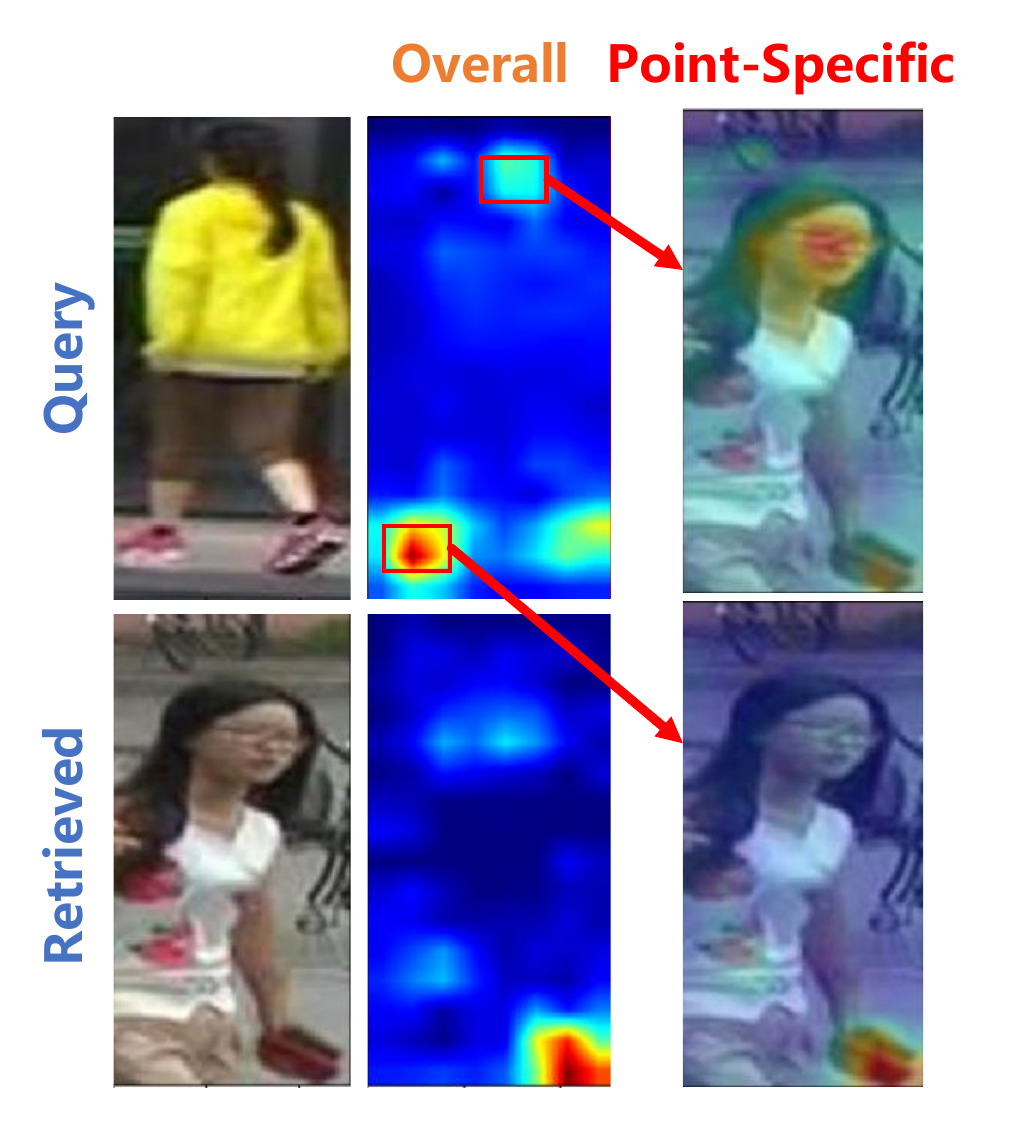}
    \caption{Explanation of the failure case by the point-specific activation map.}
    \label{fig:fail}
  \end{minipage}
\end{figure*}

Verification applications like face recognition and person re-id usually retrieve the most similar images to the query image. \textit{However, we may be more interested in a specific RoI (region of interest) than the overall image for certain circumstances, for example finding people with a similar bag or pair of shoes (i.e. RoI) in surveillance applications.}
Since our framework provides the point-to-point activation map, it can be a reasonable tool for measuring partial similarity to serve this purpose. 
With the partial similarity, we are able to \textit{interactively} retrieve images with only parts/regions similar to the query image instead of the \textit{overall} most similar images. 
\revise{As shown in Fig. \ref{fig:8}, the point-specific activation generated by the proposed method works well on retrieving people with similar clothes and faces with similar beard.}
\textbf{No explicit supervision} is adopted except the pre-trained retrieval model.

Specifically, we follow the pipeline of recent approaches on face recognition \cite{Arcface} and person re-id \cite{Re-ID}. For re-id, the model is trained and evaluated on Market-
1501 \cite{Market} where some validation images only contain a small part of a person. 
Although the model is trained with Euclidean distance without L2 normalization, cosine similarity still works well as the evaluation metric. For face recognition, we take the trained model on CASIA-WebFace \cite{CASIA} and evaluate it on FIW \cite{FIW} where the face identification and kinship relationship are available.

The interactive search is conducted by simply matching the equivalent partial feature in Eq. \ref{eq:3} ($W^{q}_{i,j}A^{q}_{i,j}$) with the reference embedding features. We first compute $W^{q}_{i,j}A^{q}_{i,j}$ as the feature of each position in the last convolutional layer, and a bilinear interpolation is adopted to generate the feature for every pixel of the original image. For a point of interest $(i,j)$, we compute  the cosine similarity between the calculated feature on $(i,j)$ and the embedding feature of the reference images as the point-specific similarity (note that the equivalent feature is normalized with the $l_{2}$ norm of the original embedding as $|E^{q}|$). In this case, the embedding features of the reference dataset do not have to be recomputed. This similarity can be also considered as 
the summation of the values in the point-specific map corresponding to $(i,j)$. In the case of Eq. \ref{eq:2} (CNN+GAP), the similarity can be simplified as $\sum_{x,y}(\sum_{k}A^{q}_{i,j,k} A^{r}_{x,y,k})/|E^{q}||E^{r}|$ ($E^{q}$ and $E^{r}$ denote the original embedding features of the query and retrieved images). \revise{If the ROI is very large, we can merge multiple equivalent features in the ROI to generate better result. Since all the objects have similar size in the face and re-id datasets, we simply adopt the equivalent feature of the center pixel in the ROI.}

Apparently, the images retrieved by the original feature are likely to be from the same identity as the query image, because the models are trained for identification task. However, the proposed interactive retrieval is able to retrieve different images based on the same query image with different RoIs as shown in Fig. \ref{fig:8}. To further quantitatively evaluate the advantage of the proposed point-specific map, we conduct human evaluation on the overall retrieval (based on the original feature) vs. the interactive retrieval (based on the point-specific map). 
Similar to Section \ref{sec:human}, the AMT workers are asked to decide which retrieval result is better given a specific RoI (see Fig. \ref{fig:human_face_interface}). We conduct experiments on both face verification and person re-id tasks. 360 evaluations of comparison pairs in total are collected from at least 40 different AMT workers. As shown in Fig. \ref{fig:human_face_reid}, the interactive retrieval holds considerably more supportive evaluations than the overall retrieval ($57.5\%$ vs. $28.0\%$ on face, and $47.5\%$ vs. $31.2\%$ on person re-id), which demonstrates the effectiveness of the proposed point-specific activation map. 

However, we do find some failure cases as shown in the last row of Fig. \ref{fig:8} with the blue bounding box. There is no shoes in the failure image, but why is this image retrieved with a top rank (i.e. $6$th rank)? We provide an in-depth analysis from the perspectives of both overall and point-specific activation maps in Fig. \ref{fig:fail}.
In the overall activation map, three regions of the query image have a high activation on the retrieved image. 
From the point-specific activation maps, there is a high activation on the purse in the retrieved image corresponding to the shoes in the query image. This might be because of their similar red color and the arm of the retrieved person may appear like a leg from a specific viewpoint. This example also validates the importance of the point-specific activation map generated by our framework for explanation.

\begin{figure}[!htbp]
\centering
\includegraphics[width=0.99\linewidth]{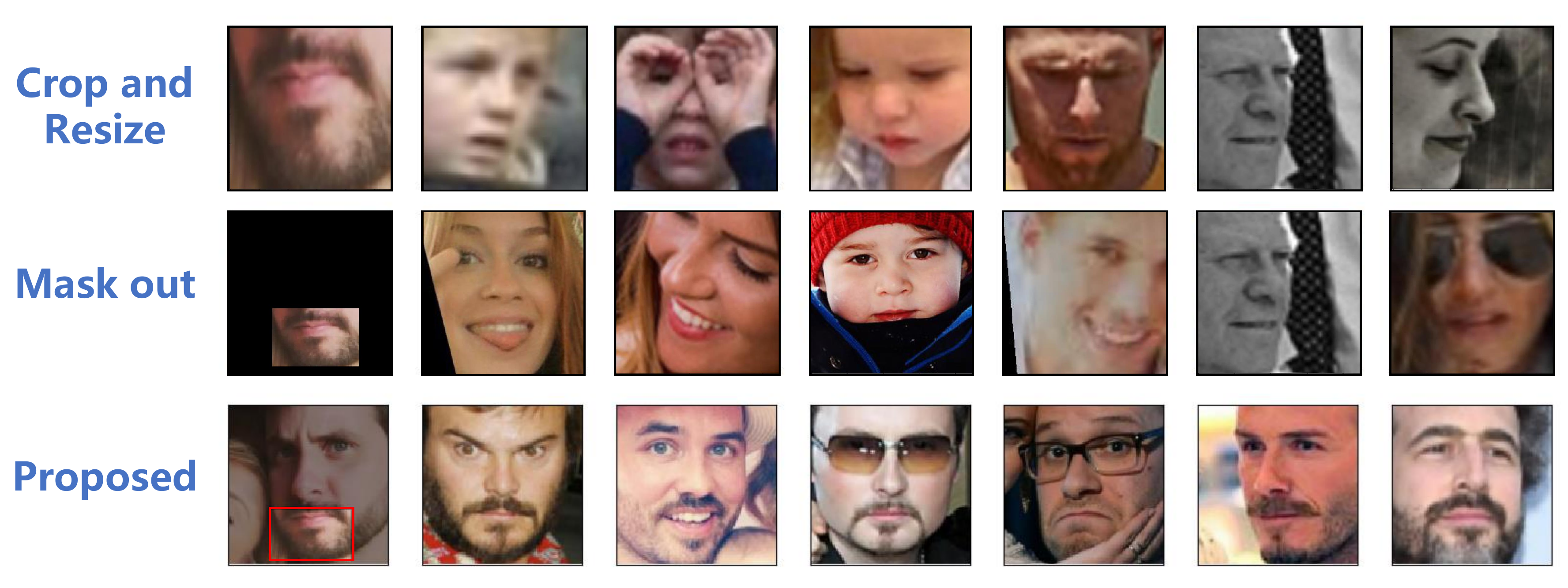}
\caption{Comparison between our method and cropping/masking for interactive retrieval on face verification. Query image is in the first column.}
\label{fig:crop_face}
\end{figure}

\begin{figure}[!htbp]
\centering
\includegraphics[width=0.99\linewidth]{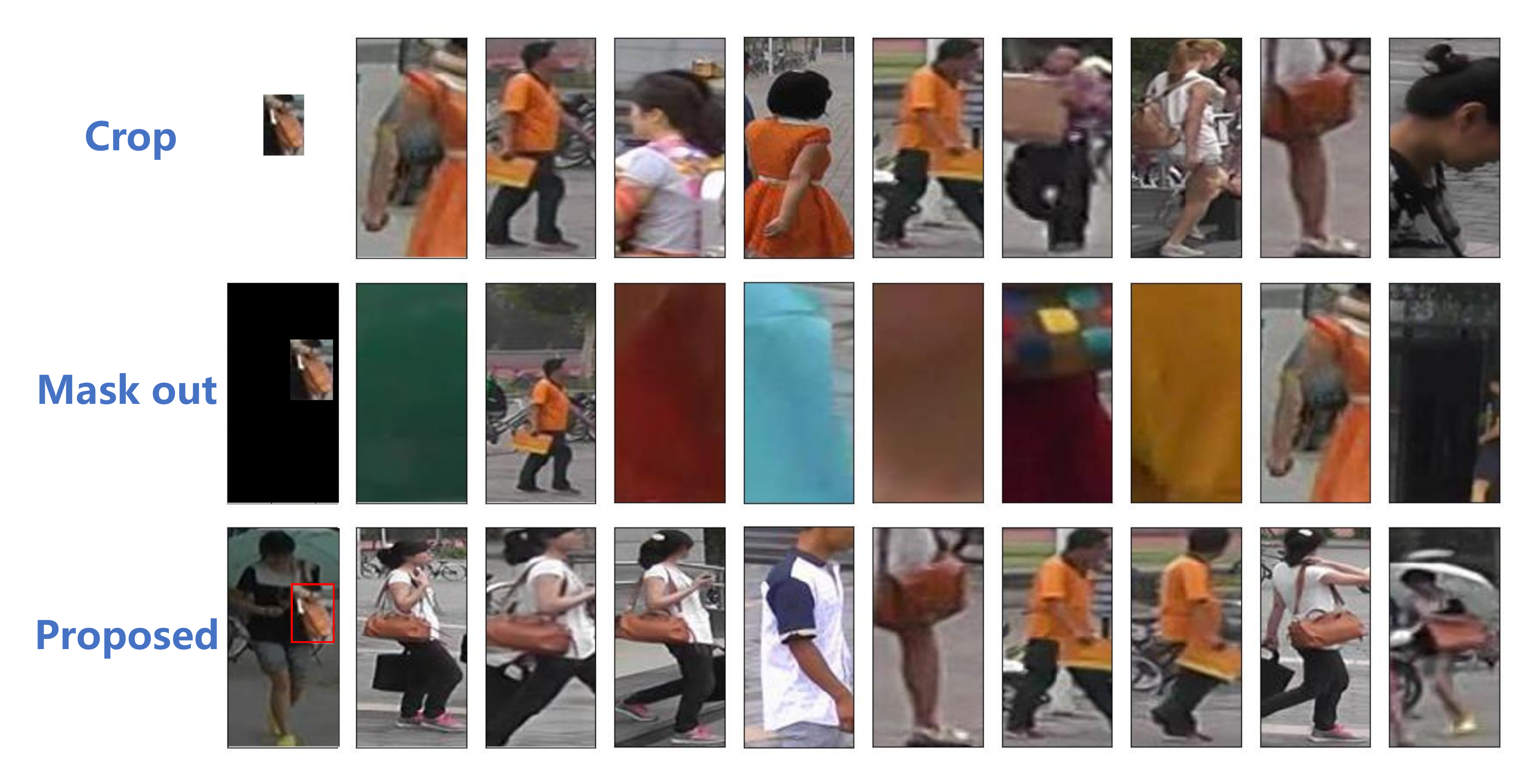}
\caption{Comparison between our method and cropping/masking for interactive retrieval on person re-identification. Query image is in the first column.}
\label{fig:crop_reid}
\end{figure}

\textbf{Comparison with Cropping/Masking.} One may argue that another possible way to achieve interactive retrieval is to crop or mask out a specific RoI. Then we can generate the feature with the specific part of image and use this feature as query to retrieve images. However, generally speaking, this operation will cause the feature to be very different from features of the original images, thus would not generate satisfactory retrieval results. And we validate this by providing the qualitative comparison of retrieval results in Figs. \ref{fig:crop_face} and \ref{fig:crop_reid}.

For face verification, flattened features followed by FC layers are usually used, because the input image size is relatively small and flattened features reserve all the information in the last convolutional layer. However, this architecture does not allow change in the input size, so one may need to resize the image to the original size after cropping a specific RoI. Another way is to assign $0$ for all pixels outside the RoI. The comparison in Fig. \ref{fig:crop_face} shows that cropping with resizing or masking are not able to generate satisfactory results.

For person re-identification, GAP is usually used so that we do not need to resize the cropped image. The results in Fig. \ref{fig:crop_reid} also show the superiority of the proposed method over cropping/masking.

\section{Conclusion}
\label{sec:conclusion}
We propose a simple yet effective framework for visual explanation of deep metric learning based on the idea of activation decomposition. The framework is applicable to a host of applications, e.g. image retrieval, face recognition, person re-id, geo-localization, etc. Experiments show the importance of visual explanation for metric learning as well as the superiority of both the overall and point-specific activation maps generated by the proposed method. Furthermore, we introduce two applications, i.e. cross-view pattern discovery and interactive retrieval, which reveal the importance of the point-specific activation map for explanation. Our work also points to interesting directions in exploring the point-specific activation maps for fine-grained information discovery and analysis.

\appendices
\section{Grad-CAM for Metric Learning}
\label{app:gradcam_metric}
For metric learning architecture discussed in Section \ref{sec:linear} of the main paper, the similarity is formulated as $S=\frac{E^{q}\cdot E^{r}}{|E^{q}||E^{r}|}$, where $E^{q}\in \R^{l}$ and $E^{r}\in \R^{l}$ are the embedding vectors of the query and retrieved images. $|x|$ denotes the L2 norm and $a\cdot b$ is the inner product of $a$ and $b$. The Grad-CAM map of the query image is given by:
\begin{equation}
\label{eq:9}
GradCAM_{i,j} = GAP\left(\frac{\partial S}{\partial A^{q}}\right) A^{q}_{i,j} 
\end{equation}
With the gradient chain rule, the gradient is written as:
\begin{equation}
\footnotesize
\label{eq:10}
\begin{aligned}
    \frac{\partial S}{\partial A^{q}} &
    = \frac{\partial \frac{(E^{r})^{T}E^{q}}{|E^{q}||E^{r}|}}{\partial A^{q}}
    = \left(\frac{E^{r}}{|E^{r}|}\right)^{T}\frac{\partial (E^{q}/|E^{q}|)}{\partial E^{q}}\frac{\partial E^{q}}{\partial A^{q}} \\
   & =\left(\frac{E^{r}}{|E^{r}|}\right)^{T}\frac{\partial (E^{q}/|E^{q}|)}{\partial E^{q}}W^{q}
    \end{aligned}
\end{equation}
By expanding $E^{r}$ as $\sum_{x,y}W^{r}_{x,y}A^{r}_{x,y}+B^{r}$ (Section \ref{sec:linear})
and merging Eq. \ref{eq:9} with Eq. \ref{eq:10}, the Grad-CAM map is reformulated as:
\begin{equation}
\scriptsize
\label{eq:11}
\begin{aligned}
    GradCAM_{i,j} & = GAP\left(\left(\frac{E^{r}}{|E^{r}|}\right)^{T}\frac{\partial (E^{q}/|E^{q}|)}{\partial E^{q}}W^{q}\right)A^{q}_{i,j}\\
    &= \frac{1}{Z}\left(\sum_{i^{*},j^{*}}(E^{r})^{T}\frac{\partial (E^{q}/|E^{q}|)}{\partial E^{q}}W_{i^{*},j^{*}}^{q}\right)A^{q}_{i,j}\\
    &=\frac{1}{Z}(E^{r})^{T}\frac{\partial (E^{q}/|E^{q}|)}{\partial E^{q}}\left(\sum_{i^{*},j^{*}}W_{i^{*},j^{*}}^{q}\right)A^{q}_{i,j}\\
    &=\frac{1}{Z}\left(\sum_{x,y}W^{r}_{x,y}A^{r}_{x,y}+B^{r}\right)
     \cdot \left(\frac{\partial (E^{q}/|E^{q}|)}{\partial E^{q}}GAP(W^{q})A^{q}_{i,j}\right)\\
    &= \frac{1}{Z}\left(\frac{\partial (E^{q}/|E^{q}|)}{\partial E^{q}}GAP(W^{q})A^{q}_{i,j}\right)\cdot\left(\sum_{x,y}W^{r}_{x,y}A^{r}_{x,y}+B^{r}\right)
    \end{aligned}
\end{equation}
Here the $Z$ is the normalization term for simplicity. $\frac{\partial E/|E|}{\partial E}$ is the $l\times l$ Jacobian matrix given by:
\begin{equation}
\small
    \frac{\partial (E/|E|)}{\partial E} = \Big(\frac{\partial (E_{i}/|E|)}{\partial E_{j}}\Big)_{i,j}= \left\{
    \begin{aligned}
     \frac{1}{|E|}\left(1-\frac{E_{i}^{2}}{|E|^{2}}\right)   & & i=j\\
    -\frac{E_{i}E_{j}}{|E|^{3}}    & & i\neq j
    \end{aligned}
    \right.
\end{equation}
$\frac{1}{|E|}$ is the normalization term. The  $\frac{\partial E/|E|}{\partial E}$ term can be removed, if we compute the gradient from $E^{q}\cdot E^{r}$ instead of $\frac{E^{q}\cdot E^{r}}{|E^{q}||E^{r}|}$.  For the dominant channel $i$, the weight $(1-\frac{E^{2}_{i}}{|E|^{2}})$ is small resulting in a more scattered activation map.

\section{Qualitative Results}
\begin{figure*}[htbp]
    \centering
    \includegraphics[width=0.95\linewidth]{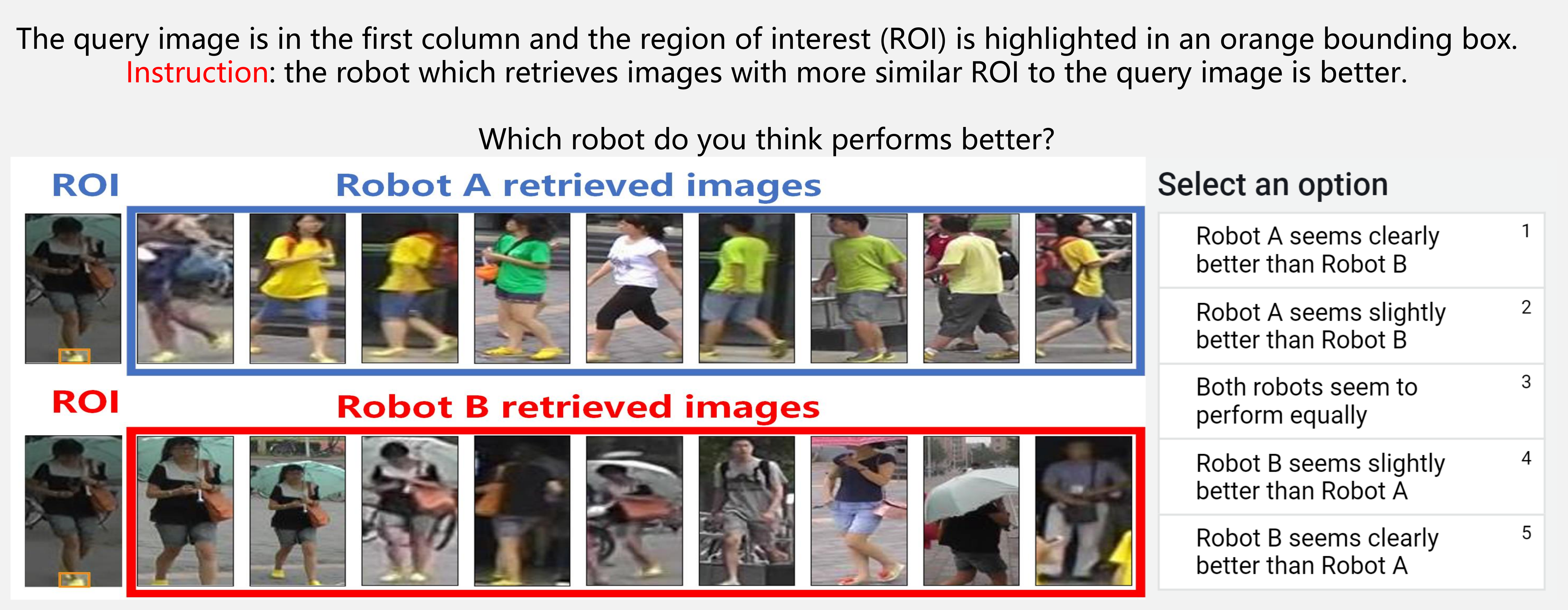}
    \caption{\revise{Human evaluation interface for interactive retrieval on person re-identification.}}
    \label{fig:human_reid_interface}
\end{figure*}

\begin{figure*}[!htbp]
\centering
\includegraphics[width=0.99\linewidth]{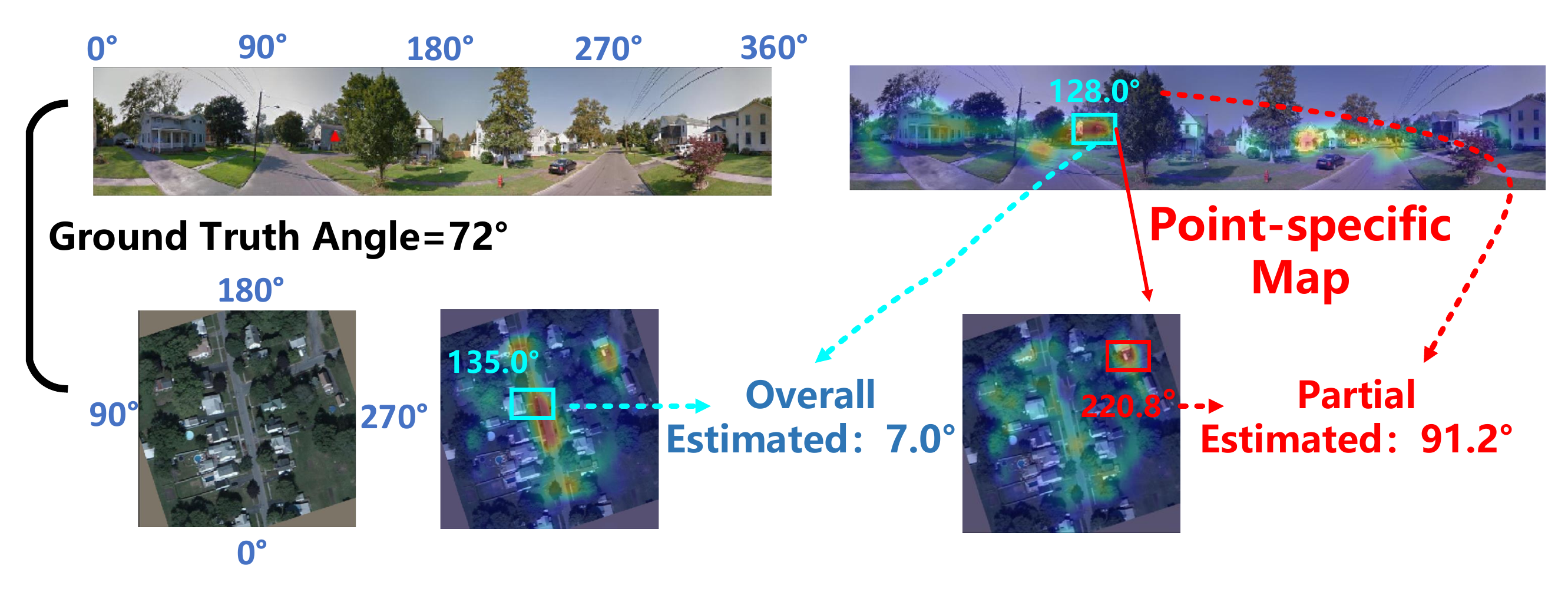}
\includegraphics[width=0.99\linewidth]{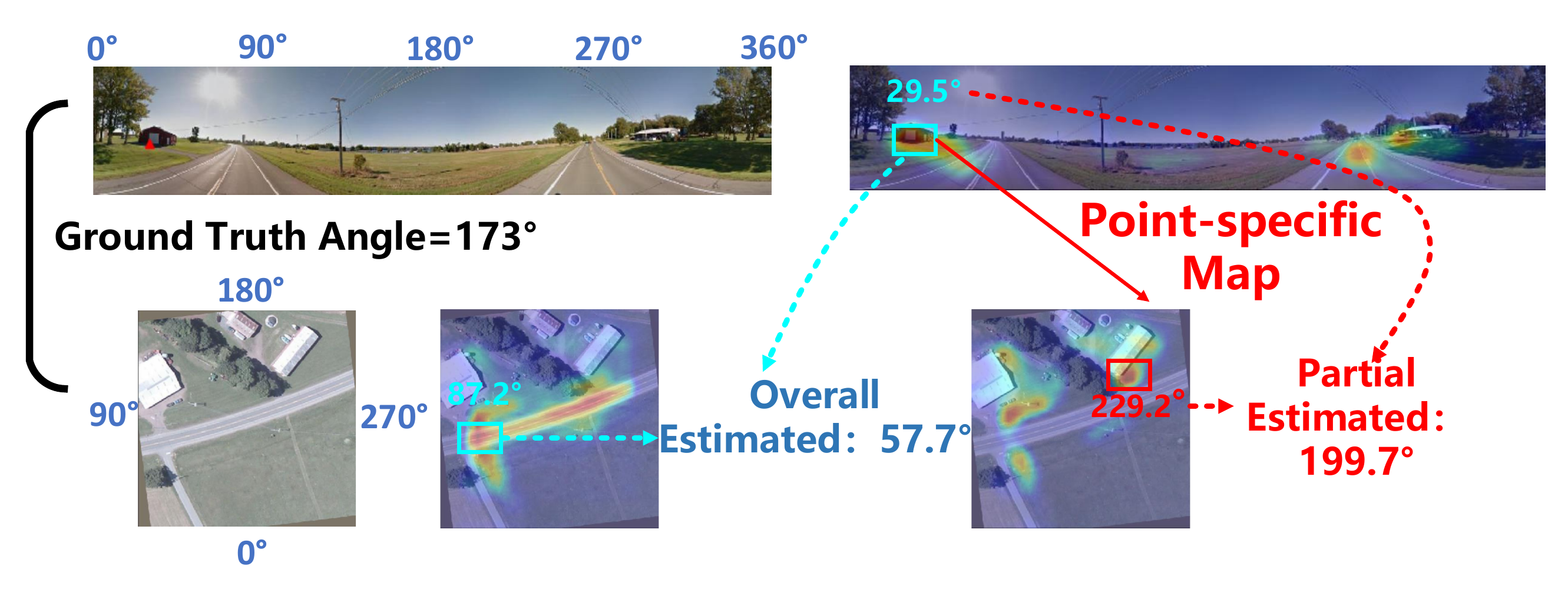}
\caption{\revise{Examples of cross-view pattern discovery, i.e., image orientation estimation. (Best viewed in color)}}
\label{fig:extra_geo}
\end{figure*}

\begin{figure*}[!htbp]
\centering
\includegraphics[width=0.99\linewidth]{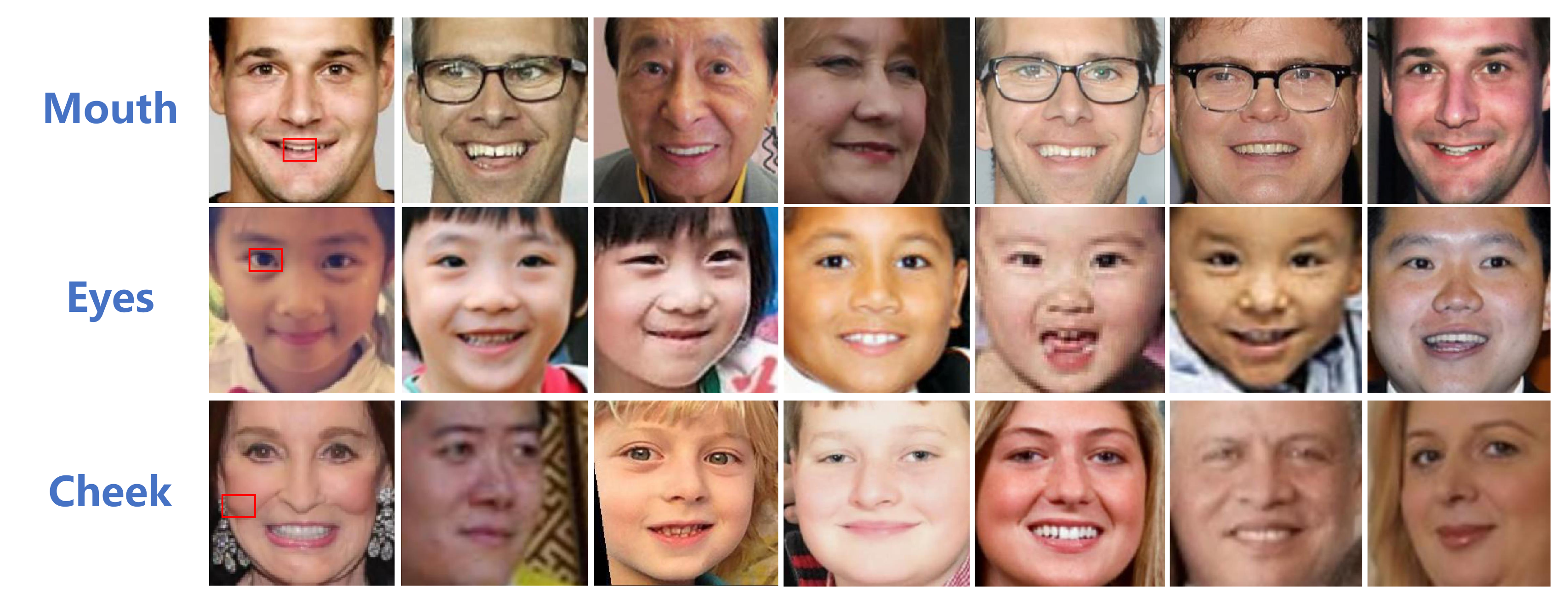}
\caption{\revise{Top retrieved images by interactive retrieval on face recognition. Red boxes on the query images (first column) highlight the Region of Interest (RoI).}}
\label{fig:extra_face}
\end{figure*}

\begin{figure*}[!htbp]
\centering
\includegraphics[width=0.99\linewidth]{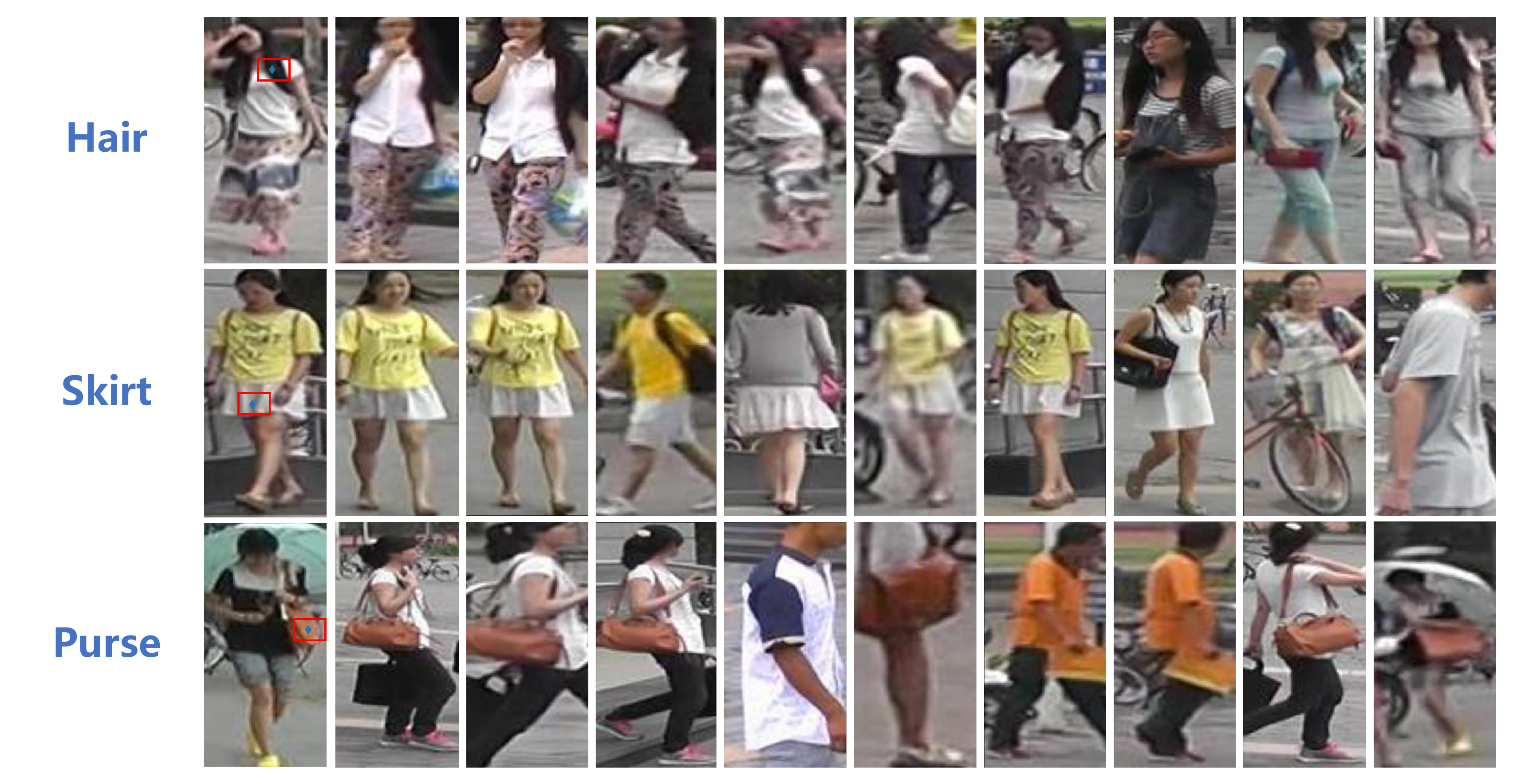}
\caption{\revise{Top retrieved images by interactive retrieval on person re-identification. Red boxes on the query images (first column) highlight the Region of Interest (RoI).}}
\label{fig:extra_reid}
\end{figure*}

\revise{In Fig. \ref{fig:human_reid_interface}, we show an example of human evaluation interface for person re-identification.} \revise{We also provide extra qualitative results for cross-view pattern discovery, interactive retrieval (both face verification and person re-identification) in Figs. \ref{fig:extra_geo}, \ref{fig:extra_face}, \ref{fig:extra_reid}.}

\ifCLASSOPTIONcaptionsoff
  \newpage
\fi



%

\bibliographystyle{IEEEtran}
\bibliography{IEEEabrv,egbib}


%

\end{document}